\documentclass{article}
     \PassOptionsToPackage{numbers, compress}{natbib}

\usepackage[final]{neurips_2022}

\usepackage[utf8]{inputenc} 
\usepackage[T1]{fontenc}    
\usepackage{hyperref}       
\usepackage{url}            
\usepackage{booktabs}       
\usepackage{amsfonts}       
\usepackage{nicefrac}       
\usepackage{microtype}      
\usepackage{xcolor}         
\usepackage{amsmath}
\usepackage{graphicx}
\usepackage{subfig}
\usepackage{multirow}
\allowdisplaybreaks[4]
\usepackage{wrapfig}

\usepackage{listings}

\definecolor{dkgreen}{rgb}{0,0.6,0}
\definecolor{gray}{rgb}{0.5,0.5,0.5}
\definecolor{mauve}{rgb}{0.58,0,0.82}

\title{Deep Fourier Up-Sampling}

\author{Man Zhou$^{1,2}$\footnotemark[1], Hu Yu$^{2}$\footnotemark[1], Jie Huang$^{2}$, Feng Zhao$^{2}$, \\ \textbf{Jinwei Gu}$^{6}$, \textbf{Chen Change Loy}$^{3}$, \textbf{Deyu Meng}$^{4,5}$, \textbf{Chongyi Li}$^{3}$\footnotemark[2]\\ 
{\tt\small$^{1}$Hefei Institute of Physical Science, Chinese Academy of Sciences, China}\\
{\tt\small$^{2}$University of Science and Technology of China, China}\\
{\tt\small$^{3}$S-Lab, Nanyang Technological University, Singapore}\\
{\tt\small$^{4}$Xi’an Jiaotong University, China}\\
{\tt\small$^{5}$Pazhou Laboratory (Huangpu), China}\\
{\tt\small$^{6}$SenseBrain Technology (SenseTime Research USA), USA}\\
{\tt\small \{manman,yuhu520,hj0117\}@mail.ustc.edu.cn, fzhao956@ustc.edu.cn},\\ 
{\tt\small gujinwei@sensebrain.site, dymeng@mail.xjtu.edu.cn}\\ {\tt\small\{ccloy,chongyi.li\}@ntu.edu.sg}\\
 \vspace{5mm}
{\tt\small \url{https://li-chongyi.github.io/FourierUp_files/}}
}

\begin{document}

\maketitle
\footnotetext[1]{Man Zhou and Hu Yu contribute equally.}
\footnotetext[2]{Corresponding author: Chongyi Li.}

\begin{abstract}
Existing convolutional neural networks widely adopt spatial down-/up-sampling for multi-scale modeling. However, spatial up-sampling operators (\emph{e.g.}, interpolation, transposed convolution, and un-pooling) heavily depend on local pixel attention, incapably exploring the global dependency. In contrast,  the Fourier domain obeys the nature of global modeling according to the spectral convolution theorem. Unlike the spatial domain that performs up-sampling with the property of local similarity, up-sampling in the Fourier domain is more challenging as it does not follow such a local property. In this study, we propose a theoretically sound Deep Fourier Up-Sampling (FourierUp) to solve these issues. We revisit the relationships between spatial and Fourier domains and reveal the transform rules on the features of different resolutions in the Fourier domain, which provide key insights for FourierUp's designs. FourierUp as a generic operator consists of three key components: 2D discrete Fourier transform,  Fourier dimension increase rules, and 2D inverse Fourier transform, which can be directly integrated with existing networks. Extensive experiments across multiple computer vision tasks, including object detection, image segmentation, image de-raining, image dehazing, and guided image super-resolution, demonstrate the consistent performance gains obtained by introducing our FourierUp. 
\end{abstract}

\section{Introduction}
Spatial down-/up-sampling has been widely used in convolutional neural networks for multi-scale modeling. For example, U-Net \cite{ronneberger2015u}, a variation of encoder-decoder, employs pooling layers to reduce the feature resolution in the encoder and then recovers the resolution using up-sampling operations in the decoder. In addition,  the feature pyramid \cite{lin2017feature,kim2018parallel,seferbekov2018feature,zhu2018bidirectional} and image pyramid \cite{pang2019efficient,liu2020ipg,adelson1984pyramid,luo2020lightweight} driven multi-scale neural networks rely on the down-/up-sampling operation to obtain multi-scale property and improve modeling capability. However,  spatial up-sampling operators (\emph{e.g.}, interpolation, transposed convolution, and un-pooling) heavily depend on local pixel attention, and thus cannot explore the global dependency that is indispensable for many computer vision tasks \cite{9156921,8767931,ronneberger2015u,7485869,fu2021unfolding,9662053,9858176,Zhou_2022_CVPR,9661815,9499114,Zhou_2021_CVPR,zhou2022memoryaugmented}. According to the spectral convolution theorem, the Fourier domain obeys the nature of global modeling, providing an alternative solution for multi-scale modeling. However, unlike the spatial domain with local similarity property, up-sampling in the Fourier domain is more challenging as it does not follow such a local property. The observation encourages us to explore deep Fourier up-sampling.

Recent studies have explored information interaction in both spatial and Fourier domains. FFC \cite{NEURIPS2020_2fd5d41e}, for instance, replaces the conventional convolution with a spatial-Fourier interaction, which consists of a spatial (or local) path that performs conventional convolution on a portion of input feature channels and a spectral (or global) path that operates in the Fourier domain. DFT \cite{mao2021deep} devises a Residual Fast Fourier Transform Block to integrate both low- and high-frequency residual information by performing the interaction between a regular spatial residual stream and a channel-wise Fourier transform stream. However, the aforementioned methodologies only interact at a single resolution scale, and the spatial-Fourier interaction potential of multiple scales in the Fourier domain has not been investigated. The key to solving this problem lies in how to implement deep Fourier up-sampling for multi-scale Fourier pattern modeling.

\begin{figure}[t]
	\centering
	\includegraphics[width=0.94\textwidth]{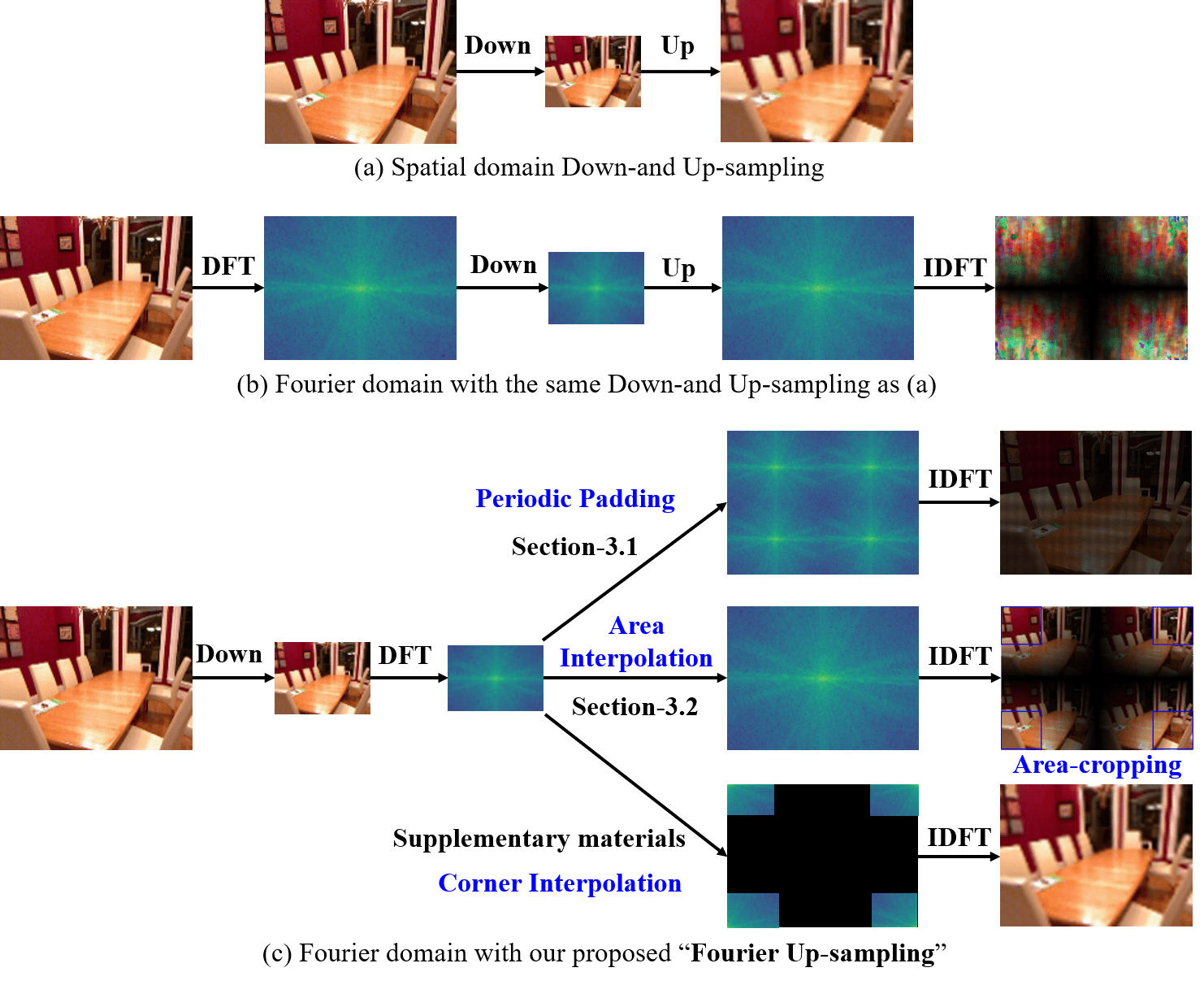}
	\caption{\textbf{Motivation.} (a) and (b) depict that arbitrary up-sampling, \emph{e.g.}, interpolation, in the Fourier domain produces sub-optimal result as it does not follow the same local similarity property as that in the spatial domain. This motivates us to design a more ingenious ``Fourier Up-Sampling'' operator, dubbed as FourierUp. It has three alternative variants: Periodic Padding, Area Interpolation/Cropping and Corner Interpolation, as illustrated in (c).}
	\label{fig:graphagg}
\end{figure}

\textbf{Challenges.} Owing to the local similarity and cross-scale position invariant properties of the spatial domain, the various spatial up-sampling operations including transposed convolution, un-pooling, and interpolation techniques are capable of using the pixel neighboring relationship to interpolate the unknown pixel values at local regions,  increasing the spatial resolution of the features, as shown in Figure \ref{fig:graphagg}(a). In contrast to the spatial domain, the Fourier domain does not share the same scale-invariant property and local texture similarity, and hence cannot implement up-sampling using the same techniques as the spatial domain, as illustrated in Figure \ref{fig:graphagg}(b).

\textbf{Solutions.}  In this paper, we wish to investigate the possibility of devising a reliable up-sampling in the Fourier domain in a theoretical sound manner.
To answer this question,  we first revisit the relationship between spatial and Fourier domains, revealing the transform rules on the features of different resolutions in the Fourier domain (see Section \ref{pp} and Section \ref{uc}). On the basis of the above rules, we propose a theoretically feasible Deep Fourier Up-Sampling (FourierUp). Specifically, we  develop three variants (Periodic Padding, Area Interpolation/Cropping and Corner Interpolation) of FourierUp (see Section \ref{AD}), as illustrated in Figure \ref{fig:graphagg}(c). Each variant consists of three key components: 2D discrete Fourier transform,  Fourier dimension increase rules, and 2D inverse Fourier transform. FourierUp is a generic operator that can be directly integrated with existing networks. Extensive experiments on multiple computer vision tasks, including object detection, image segmentation, image de-raining, image dehazing, and guided image super-resolution, demonstrate the consistent performance gains obtained by introducing our FourierUp. \textit{We believe that the proposed FourierUp could refresh the neural network designs where the spatial and Fourier information interaction at only a single resolution scale are mainstream choices.}

\textbf{Contributions.}
1) We propose Deep Fourier Up-Sampling, a novel method that enables the integration of the features of different resolutions in the Fourier domain. This is the first thorough effort to explore the Fourier up-sampling for multi-scale modeling.
2) The proposed FourierUp is a generic operator that can be directly integrated with the existing networks in a plug-and-play manner. 
3) Equipped with the  theoretically sound FourierUp, we show that existing networks could achieve consistent performance improvement across multiple computer vision tasks.

\section{Related Work}

\textbf{Spatial Up-Sampling.} Convolutional neural networks with spatial down-/up-sampling have become the \textit{de facto} structures in many computer vision tasks \cite{zhang2021aligned, zhang2021accurate, fan2020neural, ren2021deblurring, ren2018deep,zhang2018density,zhang2019image,pan2018learning,NEURIPS2019_6395ebd0}. Typically, U-Net \cite{ronneberger2015u} builds multi-scale feature maps using the encoder with down-sampling and then utilizes the up-sampling operation  to fuse the multi-scale features in the decoder. Additionally, the feature pyramid \cite{lin2017feature,kim2018parallel,seferbekov2018feature,zhu2018bidirectional} and image pyramid\cite{pang2019efficient,liu2020ipg,adelson1984pyramid,luo2020lightweight} are commonly used to obtain the multi-scale property in neural networks \cite{pang2019efficient,liu2020ipg,adelson1984pyramid,luo2020lightweight}.
Among them, spatial up-sampling plays a significant role in multi-scale modeling.  
However, existing up-sampling operations only work in the spatial domain and current studies rarely explore the potential (\emph{e.g.}, the global modeling capability) of up-sampling in the frequency domain.

\textbf{Spatial-Fourier Interaction.} Recently, several studies attempt to employ Fourier transform in deep models \cite{yang2020fda,li2020falcon,lee2018single,ding2017circnn,NEURIPS2020_2fd5d41e}. Some of these efforts use discrete Fourier transform to transfer the spatial features to the Fourier domain and then use frequency information to improve the performance of particular tasks \cite{yang2020fda,lee2018single}. Another line is to use convolution theorem to speed up the models, such as using fast Fourier transform (FFT) \cite{ding2017circnn,NEURIPS2020_2fd5d41e}. For example, FFC \cite{NEURIPS2020_2fd5d41e} replaces the convolution with the spatial-Fourier interaction. The work proposed in \cite{rippel2015spectral} uses spectral pooling to reduce feature resolution by truncating the frequency domain representation. However, all the techniques only interact with each other at a single spatial resolution and have not explored the interaction potential at multiple resolutions in both spatial and frequency domains as performing the frequency up-sampling is non-trivial.
As a tentative exploration, we study the relationship between the spatial domain and Fourier domain and reveal the transform rules over the feature of different resolutions in the Fourier domain. This delivers the underlying insights for the designs of multi-scale Fourier modeling patterns, which has the potential of versatility for different network architectures.

\section{Deep Fourier Up-Sampling}
We first explore the mapping relationship between the spatial and Fourier domains, and then present three Deep Fourier up-sampling variants, including i) periodic padding of magnitude and phase, ii) area up-sampling of magnitude and phase, and iii) corner interpolation of magnitude and phase, based on the explored transform rules. In terms of the first two variants, we provide two theorems and their proofs as follows while the third is reported in supplementary materials. 

\textbf{Definitions.} $f(x,y)\in \mathbb{R}^{2M\times 2N}$ is the $2$-times zero-inserted up-sampled version of $g(x,y) \in \mathbb{R}^{M\times N}$ in spatial domain, and $F(u,v) \in \mathbb{R}^{2M\times 2N}$, $G(u,v) \in \mathbb{R}^{M\times N}$ denote their Fourier transforms.  $H(u,v) \in \mathbb{R}^{2M\times 2N}$ is the $2$-times area-interpolation up-sampled Fourier transform of $G(u,v)$, and $h(x,y) \in \mathbb{R}^{M\times N}$ denotes their inverse Fourier transform.

\textbf{Theorem-1.} $F(u, v) = F(u+M, v) = F(u, v+N) = F(u+M, v+N)$ and $G(u,v) = \frac{F(u,v)}{4}$ where $u=0, 1, 2, \dots, N-1$ and $v = 0, 1, 2, \dots, M-1$. $F(u,v)$ is exactly the periodic padding of $G(u,v)$ where $G(u,v)$ is exactly the quarter of $F(u,v)$ with the value being $\frac{1}{4}$ times decay.

\textbf{Theorem-2.} $H(2u, 2v) = H(2u+1, 2v) = H(2u, 2v+1) = H(2u+1, 2v+1) =G(u,v)$ with $u=0,1,\dots,M-1$ and $v=0,1,\dots,N-1$ and \begin{equation} \label{gif}
		\begin{aligned}
          h(x,y) &= \frac{A(x,y)}{4} g(x,y) \\
          h(x+M,y) &= \frac{A(2M-x,y)}{4} g(x,y) \\
          h(x,y+N) &= \frac{A(x,2N-y)}{4} g(x,y)  \\
          h(x+M,y+N) &= \frac{A(2M-x,2N-y)}{4} g(x,y)
  \end{aligned}
\end{equation}
where $A(x,y)=1 + e^{\frac{j\pi x}{M}}+ e^{\frac{j\pi y}{N}} + e^{j\pi(\frac{x}{M}+\frac{y}{N})}$ and $x=0, 1, \dots M-1, y=0, 1, \dots N-1$.

\textbf{Theorem-3.} Suppose the corner interpolated $F^{cor}_{G}(u,v)$ of the Fourier map $G(u,v) \in \mathbb{R}^{M\times N}$, it holds that the inverse Fourier transform $f^{cor}_{g}(x,y)$ of $F^{cor}_{G}(u,v)$
\begin{equation}
	\small
		\begin{aligned}
          f^{cor}_{g}(x, y) = g(\frac{x'}{2},\frac{y'}{2})e^{j\pi (\frac{x'}{2} + \frac{y'}{2})}\frac{(-1)^{(x+y)}}{4},
  \end{aligned}
\end{equation}
where $x'=2x$ and $y'=2y$, $x=0,1,\dots,M-1$ and $y=0,1,\dots,N-1$.

\subsection{Proof-1 of Theorem-1: Periodic Padding of Magnitude and Phase} \label{pp}

Note that $f(x, y) \in \mathbb{R}^{2M\times 2N}$ is up-sampled over $g(x,y) \in \mathbb{R}^{M\times N}$ by a factor of $2$. The relationship between $g(x,y)$ and $f(x,y)$ can be written as
\begin{eqnarray}
\boldsymbol f(x, y) & = & \left\{\begin{array}{l}
g(\frac{x}{2}, \frac{y}{2}), \quad x=2m, y=2n \\
0, \quad\quad\quad\quad \text{others}
\end{array}\right.
\end{eqnarray}
where $m=1,2,\dots,M-1$ and $n=1,2,\dots,N-1$, the Fourier transform $F(u,v)$  of $f(x,y)$ is expressed as
\begin{equation} \label{pequ}
		\begin{aligned}
		  F(u,v) &= \frac{1}{4MN} \sum_{x=0}^{2M-1} \sum_{y=0}^{2N-1} f(x,y) e^{-j2\pi (\frac{ux}{2M} + \frac{vy}{2N})} \\
           &= \frac{1}{4MN}\sum_{x=0}^{M-1} \sum_{y=0}^{N-1} f(2x, 2y) e^{-j2\pi (\frac{u(2x)}{2M} + \frac{v(2y)}{2N})}\\ 
           &= \frac{1}{4MN}\sum_{x=0}^{M-1} \sum_{y=0}^{N-1} f(2x, 2y) e^{-j2\pi (\frac{ux}{M} + \frac{vy}{N})} \\
           &= \frac{1}{4MN}\sum_{x=0}^{M-1} \sum_{y=0}^{N-1} g(x, y) e^{-j2\pi (\frac{ux}{M} + \frac{vy}{N})}.
   \end{aligned}
\end{equation}

Then, we show the periodicity of $F(u,v) \in R^{2M \times 2N}$  with $M$ and $N$. It means $F(u, v) = F(u+M, v) = F(u, v+N) = F(u+M, v+N)$  with $u=0, 1, 2, \dots, N-1$ and $v = 0, 1, 2, \dots, M-1$. We take the $F(u, v) = F(u+M, v)$ for example and recall Eq. \eqref{pequ} as
\begin{equation}
	\small
		\begin{aligned}
          F(u+M, v) &= \frac{1}{4MN}\sum_{x=0}^{M-1} \sum_{y=0}^{N-1} f(2x, 2y) e^{-j2\pi (\frac{(u+M)x}{M} + \frac{vy}{N})} \\  
          &= \frac{1}{4MN}\sum_{x=0}^{M-1} \sum_{y=0}^{N-1} f(2x, 2y) e^{-j2\pi (\frac{(ux}{M} + \frac{vy}{N})}e^{-2j\pi x}\\  
           &= \frac{1}{4MN}\sum_{x=0}^{M-1} \sum_{y=0}^{N-1} f(2x, 2y) e^{-j2\pi (\frac{(ux}{M} + \frac{vy}{N})}\\ 
          &= F(u, v),
        \end{aligned}
\end{equation}
where $e^{-2\pi jx} = 1$ for any integer $x$. Similarly, we can proof the periodicity of $F(u,v)$ as well.

Based on the above proof, the DFT of $g(x,y)$ can be formulated as:
\begin{equation}
	\small
		\begin{aligned}
          G(u,v) &= \frac{1}{MN}\sum_{x=0}^{M-1} \sum_{y=0}^{N-1} g(x,y) e^{-j2\pi (\frac{ux}{M} + \frac{vy}{N})}.
  \end{aligned}
\end{equation} 
Revising Eq. \eqref{pequ}, we can figure out that $G(u,v) = \frac{F(u,v)}{4}$.

\subsection{Proof-2 of Theorem-2: Area Interpolation of Magnitude and Phase} \label{uc}
The 2D Inverse Discrete Fourier transform (IDFT) of $G(u,v)$ can be written as:
\begin{equation}
	\small
		\begin{aligned}
          g(x,y) = \frac{1}{MN}\sum_{u=0}^{M-1} \sum_{v=0}^{N-1} G(u,v) e^{j2\pi (\frac{ux}{M} + \frac{vy}{N})}.     
  \end{aligned}
\end{equation} 
We up-sample $G(u,v)$ with a size of $M \times N$ to get $H(u,v)$ with a size of
$2M \times 2N$. Specifically, the area interpolation shown in Figure \ref{fig:sub}(b) is used for interpolation and then the interpolated pixels are the same as the original pixel in the  $2\times 2$ local regions. Namely, $H(2u, 2v) = H(2u+1, 2v) = H(2u, 2v+1) = H(2u+1, 2v+1) =G(u,v)$ with $u=0,1,\dots,M-1$ and $v=0,1,\dots,N-1$. Similar to Eq. \eqref{pequ}, we can infer
\begin{equation} \label{hif}
	\resizebox{\linewidth}{!}{
		\begin{math}
		\begin{aligned}
		  h(x,y) &= \frac{1}{4MN}\sum_{u=0}^{2M-1} \sum_{v=0}^{2N-1} H(u,v) e^{j2\pi (\frac{ux}{2M} + \frac{vy}{2N})} \\
           &= \frac{1}{4MN} \sum_{u=0}^{M-1} \sum_{v=0}^{N-1} H(2u, 2v) e^{j2\pi (\frac{2ux}{2M} + \frac{2vy}{2N})} + \frac{1}{4MN}
           \sum_{u=0}^{M-1} \sum_{v=0}^{N-1} H(2u+1, 2v) e^{j2\pi (\frac{(2u+1)x}{2M} + \frac{2vy}{2N})}\\ 
           &+ \frac{1}{4MN} \sum_{u=0}^{M-1} \sum_{v=0}^{N-1} H(2u,2v+1) e^{j2\pi (\frac{2ux}{2M} + \frac{(2v+1)y}{2N})} + \frac{1}{4MN} \sum_{u=0}^{M-1} \sum_{v=0}^{N-1} H(2u+1, 2v+1) e^{j2\pi (\frac{(2u+1)x}{2M} + \frac{(2v+1)y}{2N})}\\
           &= \frac{1}{4MN} \sum_{u=0}^{M-1} \sum_{v=0}^{N-1} G(u,v) e^{j2\pi (\frac{ux}{M} + \frac{vy}{N})}[1 + e^{\frac{j\pi x}{M}}+ e^{\frac{j\pi y}{N}}+ e^{j\pi(\frac{x}{M}+\frac{y}{N})}].
   \end{aligned}
\end{math}}
\end{equation}
Similarly, we can write $g(x,y)$ as
\begin{equation} \label{gif}
		\begin{aligned}
          g(x,y) &= \frac{1}{MN}\sum_{u=0}^{M-1} \sum_{v=0}^{N-1} G(u,v) e^{j2\pi (\frac{ux}{M} + \frac{vy}{N})}.
  \end{aligned}
\end{equation}
 Recalling Eq. \eqref{hif} and Eq. \eqref{gif}, we can infer 
\begin{equation} \label{gif}
		\begin{aligned}
          h(x,y) &= \frac{1 + e^{\frac{j\pi x}{M}}+ e^{\frac{j\pi y}{N}} + e^{j\pi(\frac{x}{M}+\frac{y}{N})}}{4} g(x,y).
  \end{aligned}
\end{equation}
where $x=0, 1, \dots M-1, y=0, 1, \dots N-1$. We remark $1 + e^{\frac{j\pi x}{M}}+ e^{\frac{j\pi y}{N}} + e^{j\pi(\frac{x}{M}+\frac{y}{N})}=A(x,y)$,
\begin{equation}
	\resizebox{\linewidth}{!}{
		\begin{math}
	\begin{aligned}
    \left | A(x,y) \right |^{2} 
    &=(1 + \cos\pi\frac{x}{M} +\cos\pi\frac{y}{N} +\cos\pi(\frac{x}{M}+\frac{y}{N}))^{2} + (\sin\pi\frac{x}{M} + \sin\pi\frac{y}{N}+ \sin\pi(\frac{x}{M}+\frac{y}{N}))^{2}. \nonumber
  \end{aligned}
\end{math}}
\end{equation}
We can find that the variable $x$ shares the same operations as the variable $y$. For brevity, we only take the operation of variable $x$ as example 
\begin{equation}
	\small
	\begin{aligned}
    \frac{\partial \left | A(x,y) \right |^{2}}{\partial x} 
    &= \frac{2\pi}{M}(1 + \cos\pi\frac{x}{M} +\cos\pi\frac{y}{N} +\cos\pi(\frac{x}{M}+\frac{y}{N}))(-\sin\pi\frac{x}{M} - \sin \pi(\frac{x}{M}+\frac{y}{N})) \\
    &+ \frac{2\pi}{M}(\sin\pi\frac{x}{M} + \sin\pi\frac{y}{N}+ \sin \pi(\frac{x}{M}+\frac{y}{N}))(\cos\pi(\frac{x}{M})+\cos\pi(\frac{x}{M}+\frac{y}{N})) \\
    &= -\frac{4\pi}{M}(\sin\pi\frac{x}{M})(1+\cos\pi(\frac{y}{N})).
    \end{aligned}
\end{equation}
Equally, we have
\begin{equation}
	\small
	\begin{aligned}
    \frac{\partial \left | A(x,y) \right |^{2}}{\partial y}  &=  -\frac{4\pi}{N}\sin\pi\frac{y}{N}(1+\cos\pi(\frac{x}{M})).
    \end{aligned}
\end{equation}

\begin{wrapfigure}{l}{0.34\textwidth}
\begin{center}
\includegraphics[width=0.32\textwidth]{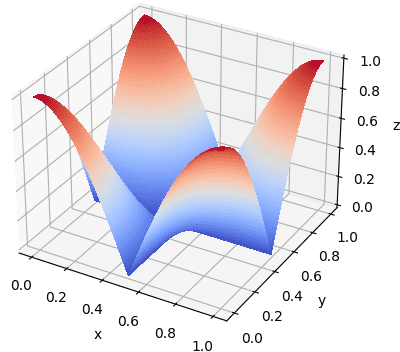} 
\end{center}
\caption{\textbf{The surface of $A(x,y)$.}} \label{csf}
\end{wrapfigure}

We prove that the partial derivative of $\left | A(x,y) \right |$ on both x and y is negative for $x \in [0, M-1 ]$ and  $y\in [0, N-1 ]$. Besides, we have
\begin{equation}
	\small
	\begin{aligned}
    \left | A(x,y) \right |  = \left | A(2M-x,y) \right |
    = \left | A(2M,2N-y) \right | 
    = \left | A(2M-x,2N-y) \right |.
    \end{aligned}
\end{equation}
That is to say, the intensity drops from the side to the center, shown in Figure \ref{csf}. Specifically, the intensity drops to zero at the position of $x=M$ or $y=N$. 

\begin{figure*}
\rule{\textwidth}{0.4pt}
\begin{minipage}[t]{\textwidth}
    \begin{lstlisting}[language={Python}]
    def  DFU_Padding(X):
     # X: input with shape [N, C, H, W]
     # A and P are the amplitude and phase 
        A, P = FFT(X)  
        
        # Fourier up-sampling transform rules 
        A_pep = Periodic-Padding(A)   
        P_pep = Periodic-Padding(P)  
        A_pep = Convs_1x1(A_pep)     
        P_pep = Convs_1x1(P_pep)     
        
        # Inverse Fourier transform
        Y = iFFT(A_pep, P_pep)      
            
        Return Y #[N, C, 2H, 2W]
\end{lstlisting}
    \end{minipage}
    \hfill
    \vline
    \begin{minipage}[t]{\textwidth}
    \begin{lstlisting}[language={Python}]
def  DFU_AreaInterpolation(X):
 # X: input with shape [N, C, H, W]
 # A and P are the amplitude and phase
    A, P = FFT(X)  
    
    # Fourier up-sampling transform rules 
    A_aip = Area-Interpolation(A) 
    P_aip = Area-Interpolation(P) 
    A_aip = Convs_1x1(A_aip) 
    P_aip = Convs_1x1(P_aip)  
    
    # Inverse Fourier transform
    Y = iFFT(A_aip, P_aip) 
    
    #Area Cropping
    Y = Area-Cropping(Y)  
    Y = Resize(Y) 
    
    Return Y #[N, C, 2H, 2W]
    \end{lstlisting}
    \end{minipage}

\rule{\textwidth}{1pt}
\caption{\textbf{Pseudo-code of the two variants of the proposed deep Fourier up-sampling.} The left is the \textit{periodic padding variant} while the right is the \textit{area interpolation-cropping variant}.}
    \label{fig:my_label}
\end{figure*}

\begin{figure}[t]
	\centering
	\includegraphics[width=\textwidth]{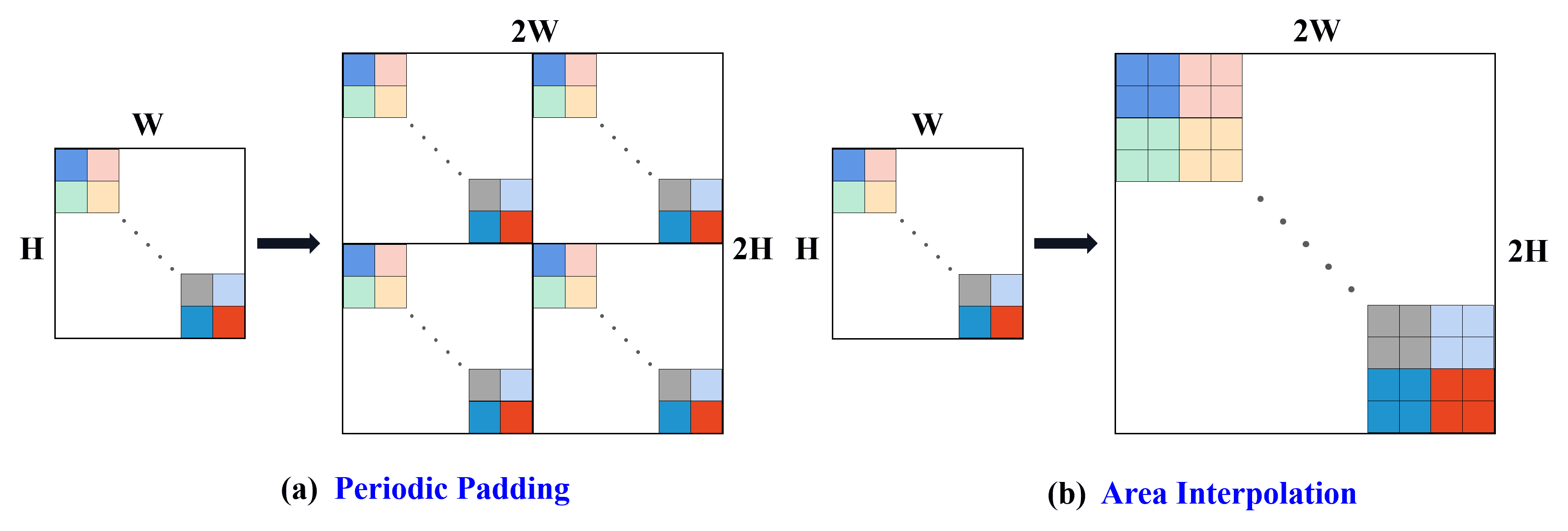}
	\caption{\textbf{The illustrations of (a) periodic padding and (b) area interpolation in Figure \ref{fig:my_label}.} Each small color square represents a pixel of the amplitude/phase component in the Fourier domain.}
	\label{fig:sub}
\end{figure}

\newpage

\subsection{Architectural Design} \label{AD}
Recall the \textbf{Theorem-1} and \textbf{Theorem-2}, we propose two deep Fourier up-sampling variants: Periodic Padding and Area Interpolation-Cropping.  

\textbf{Periodic Padding Up-Sampling.} The pseudo-code of Periodic Padding Up-Sampling is shown in the left of Figure \ref{fig:my_label}.  Given an image $\mathrm{X \in \mathbb{R}^{H \times W \times C}}$, we first adopt the Fourier transform $\mathrm{FFT(X)}$ to obtain its amplitude component $\mathrm{A}$ and phase component $\mathrm{P}$. We then perform the periodic padding over $\mathrm{A}$ and $\mathrm{P}$  two times in both the $H$ and $W$ dimensions, as illustrated in Figure \ref{fig:sub}(a). The padded $\mathrm{A\_pep}$ and $\mathrm{P\_pep}$ are then fed into two independent convolution module with $1\times 1$ kernel and followed by the inverse Fourier transform $\mathrm{iFFT(.)}$ to project the padded ones back to spatial domain.

\textbf{Area Interpolation-Cropping Up-Sampling.} The pseudo-code of Area interpolation-Cropping Up-Sampling is shown in the right  of Figure \ref{fig:my_label}. We first conduct the Area interpolation over the phase and amplitude by $2\times2$ area interpolation with the same pixel, as illustrated in Figure \ref{fig:sub}(b). We then employ the inverse Fourier transform to project the interpolated ones back to spatial domain. As described in Section \ref{uc}, the inverse spatial representation will be periodic while the pixel value will be decay. The degree of decay of the pixel increases when the pixel is closer to the center. To better maintain the information, we perform the area cropping operation (detailed in Figure \ref{fig:graphagg}) in the four corners with the $\frac{W}{2} \times \frac{H}{2}$ size and then merge them together as a whole at spatial dimension, finally resize them to the size of $2H \times 2W$.

Note that albeit being designed on the basis of strict theories, both constructed spectral up-sampling modules contain certain approximations, like a learnable $1\times 1$ convolution operator instead of strictly $1/4$ as described in Theorem-1 of main manuscript, and an approximation cropping to preserve the map corners instead of accurate $A$ mapping as proved in Theorem-2 of main manuscript  and Theorem-3 of supplementary materials. Such strategy makes the proposed modules able to be more easily implemented and more flexibly represent real data spectral structures. It is worth noting that this should be the first attempt for constructing easy equitable spectral upsampling modules, and hope it would inspire more effective and rational ones from more spectral perspectives.

\vspace{-1.2mm}
\section{Experiments}
\vspace{-1.2mm}
To demonstrate the efficacy of our proposed deep Fourier up-sampling, we conduct extensive experiments on multiple computer vision tasks, including object detection, image segmentation, image de-raining, image dehazing, and guided image super-resolution. We provide more experimental results in the supplementary material.

\subsection{Experimental Settings} 
\label{4.1section}
\textbf{Object Detection.} Following \cite{7485869}, the PASCAL VOC 2007 and 2012 training sets \cite{2010The} are used as training data. The PASCAL VOC 2007 testing set is used for evaluations as the ground truth annotations of VOC 2012 testing set are not publicly available. We employ the FPN-based Faster RCNN \cite{7485869} with ResNet50 backbone and YOLO-v3 with Darknet53 \cite{9156454} as baselines.

\textbf{Image Segmentation.} Following \cite{chen2021transunet, fu2020domain}, Synapse Dataset and CANDI Dataset are used as the testbed of medical image segmentation. We adopt the two representative image segmentation algorithms, U-Net \cite{ronneberger2015u} and Att-UNet \cite{oktay2018attention}, as the base models.

\textbf{Image De-raining.}  Following \cite{Ren_2019_CVPR},  we choose two widely-used standard benchmark datasets, including Rain200H and Rain200L, for evaluations.  we employ two representative de-raining methods,  LPNet with up-sampling \cite{8767931} and PReNet without up-sampling \cite{Ren_2019_CVPR}, as baselines.

\textbf{Image Dehazing.} Following \cite{9156921}, we employ RESIDE\cite{8451944} dataset \cite{8296874} for evaluations. We also use two different network designs AODNet \cite{Li_2017_ICCV} without up-sampling operator and MSBDN \cite{9156921} with up-sampling operator,  for validation.

\textbf{Guided Image Super-resolution.} Following  \cite{9662053,9758796}, we adopt the pan-sharpening, the representative task of guided image super-resolution, for evaluations. The WorldView II, WorldView III, and GaoFen2 in \cite{9662053,9758796} are used for evaluations. We employ  two different network designs for validation, including PANNET \cite{pannet} without up-sampling operator and DCFNET \cite{Wu_2021_ICCV} with up-sampling operator.  

Several widely-used image quality assessment (IQA) metrics are employed to evaluate the performance, including the relative dimensionless global error in synthesis (ERGAS) \cite{ergas}, the peak signal-to-noise ratio (PSNR), the spectral angle mapper (SAM) \cite{sam}, DSC, and HD95.

\begin{figure}[h!t]
	\centering
	\includegraphics[width=\textwidth]{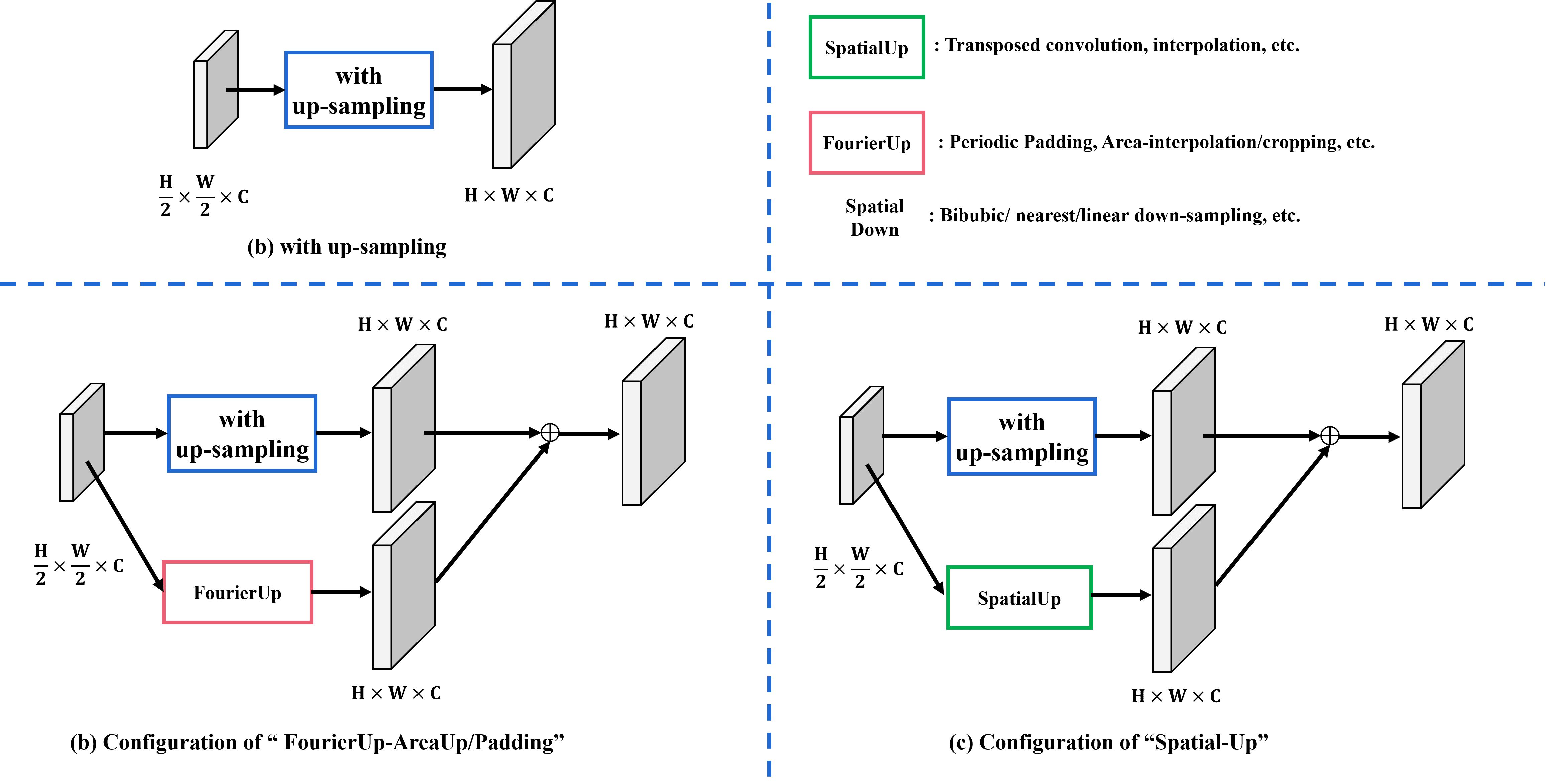}
	\caption{\textbf{The implementation details of FourierUp into the existing baselines with up-sampling.}}
	\label{fig:id}
\end{figure}

\subsection{Implementation Details} \label{4.2section}
Regarding the above competitive baselines, they can be divided into two categories: one with spatial up-sampling (\cite{ronneberger2015u}, Att-UNet \cite{oktay2018attention}, DCFNET \cite{Wu_2021_ICCV},  LPNet \cite{8767931}, MSBDN \cite{9156921}) and another one without spatial up-sampling (PReNet \cite{Ren_2019_CVPR}, AODNet\cite{Li_2017_ICCV}, PANNET \cite{pannet}). 
The purpose of the exploration on the baselines without spatial up-sampling is to show the versatility of our FourierUp for different network structures. Different from directly replacing the spatial up-sampling with the FourierUp in the baselines with spatial up-sampling, we need to encapsulate the FourierUp for the baselines without spatial up-sampling, in which a down-sampling operation is introduced to first reduce the resolution of features. We provide the detailed structures of the encapsulated FourierUp and the baselines with the FourierUp in the Figure \ref{fig:id} and supplementary material.

For the baselines with spatial up-sampling, we perform the comparison over four configurations:
\begin{itemize}
\setlength{\itemsep}{0pt}
\setlength{\parsep}{0pt}
\setlength{\parskip}{0pt}
\item [1)] \textbf{Original}: the baseline without any changes; 
\item [2)] \textbf{FourierUp-AreaUp}:  replacing the original model's spatial up-sampling  with the union of the Area-Interpolation variant of our FourierUp and the spatial up-sampling itself;
\item[3)] \textbf{FourierUp-Padding}:  replacing the original model's spatial up-sampling operator  with the union of the Periodic-Padding variant of our FourierUp and  the spatial up-sampling itself; 
\item[4)] \textbf{Spatial-Up}:   replacing the variants of  FourierUp in the settings of $2)/3)$ with the spatial up-sampling. For a  fair comparison, we use the same number of trainable parameters as $2)/3)$.
\end{itemize}

For the baselines without spatial up-sampling, we perform the comparison over four configurations:
\begin{itemize}
\setlength{\itemsep}{0pt}
\setlength{\parsep}{0pt}
\setlength{\parskip}{0pt}
\item [1)] \textbf{Original}: the baseline without any changes; 
\item [2)] \textbf{FourierUp-AreaUp}:  replacing the original model's convolution with the encapsulated FourierUp that is equipped with the Area-Interpolation variant;
\item[3)] \textbf{FourierUp-Padding}: replacing the original model's convolution  with the encapsulated FourierUp that is equipped with the Periodic-Padding variant;
\item[4)] \textbf{Spatial-Up}:   replacing the the encapsulated FourierUp of the settings of $2)/3)$ with the spatial up-sampling. For a  fair comparison, we use the same number of trainable parameters as $2)/3)$.
\end{itemize}

\subsection{Comparison and Analysis} \label{integration}

\begin{figure}[h!t]
	\centering
	\includegraphics[width=0.93\textwidth]{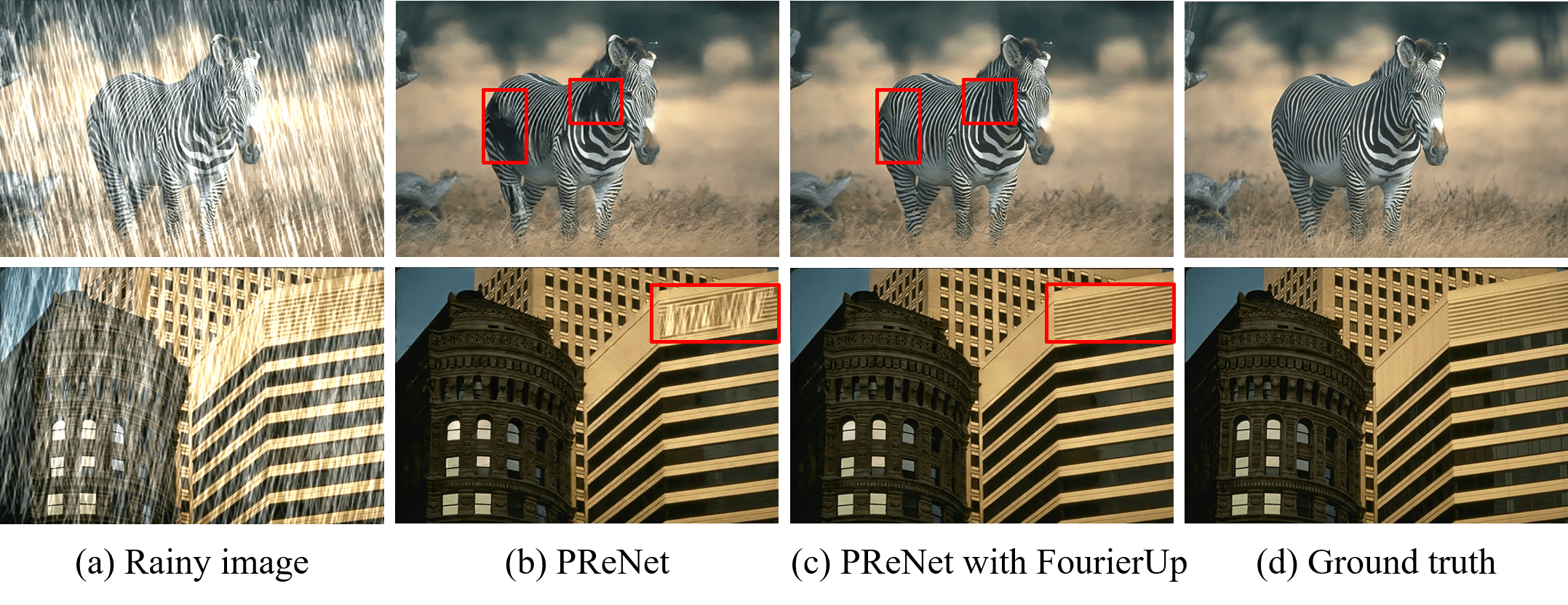}
 	\setlength{\abovecaptionskip}{-0.02cm}
	\caption{\textbf{Visual comparison of PReNet on the Rain200H.}}
	\label{fig:dr}
   	\vspace{-2.2mm}
\end{figure}

\textbf{Quantitative Comparison.} We perform the model performance comparison over different configurations, as described in implementation details. The quantitative results are presented in Tables \ref{tab:dr} to \ref{tab:ps} where the best and second best results are highlighted in bold and Underline. From the results, by integrating with our proposed two FourierUp variants, we can observe performance gain against the baselines across all the datasets in all tested tasks, suggesting the effectiveness of our approach. For example, for the PReNet of Table \ref{tab:dr}, ``FourierUp-padding'' and ``FourierUp-AreaUp'' obtain 0.83dB/0.65dB and 2.1dB/1.9dB PSNR gains than the  ``Original'', 0.52dB/0.34dB and 1.7dB/1.5dB PSNR gains than ``Spatial-Up'' on the Rain200H and Rain200L datasets, respectively. Such results validate the effectiveness of our proposed FourierUp.  The corresponding visualization consistently supports the analysis in Figure \ref{fig:dr}, where the FourierUp is capable of better maintaining the details.

\textbf{Qualitative  Comparison.} Due to the limited space, we only report the visual results of the de-raining/dehazing task in Figures \ref{fig:dr} and \ref{fig:dh} that can more clearly show the effectiveness of FourierUp.  More  results can be found in the supplementary materials. As shown,  integrating the FourierUp with the original baseline achieves more visually pleasing results. Specifically, zooming-in the red box region of Figures  \ref{fig:dr} and \ref{fig:dh}, the model equipped with the FourierUp is capable of better recovering the texture details while removing the rain/hazy effect.

\begin{table*}[!htb]
\small
\centering
\renewcommand{\arraystretch}{1.1}
\caption{\textbf{Quantitative comparisons of image de-raining.}}
\begin{tabular}{l l |c c c c }
    \hline
    \multirow{2}{*}{Model} & \multirow{2}{*}{Configurations}& \multicolumn{2}{c}{Rain200H} & \multicolumn{2}{c}{Rain200L}  \\
    &&PSNR & SSIM & PSNR & SSIM  \\
    \hline
    \multirow{4}{*}{LPNet} & Original& 22.907 & 0.775 & 32.461& 0.947  \\
     &Spatial-Up & 22.956 & 0.777 & 32.522 & 0.950  \\
    &FourierUp-AreaUp & \underline{22.163} & \underline{0.783} & \underline{32.681} & \underline{0.954}  \\
    &FourierUp-Padding & \textbf{23.295} & \textbf{0.786} & \textbf{32.835} & \textbf{0.956}  \\
    \hline
    \multirow{4}{*}{PReNet} & Original& 29.041
 & 0.891 & 37.802  & 0.981  \\
     &Spatial-Up & 29.357 & 0.901 & 38.271 & 0.985  \\
    &FourierUp-AreaUp & \underline{29.690} & \underline{0.903} & \underline{39.776} & \underline{0.985} \\
    &FourierUp-Padding & \textbf{29.871} & \textbf{0.908} & \textbf{39.971}  & \textbf{0.987}  \\
    \hline
\end{tabular}
\label{tab:dr}
\end{table*}

\begin{minipage}[c]{0.48\textwidth}
\small
\centering
\renewcommand{\arraystretch}{1.1}
\captionof{table}{\textbf{Comparison over image dehazing.}}
\begin{tabular}{c  c |c c}
    \hline
    Model                    & configurations   & PSNR   & SSIM  \\
    \hline
    
    \multirow{4}{*}{AODNet} & Original        & 18.80  & 0.834    \\
                             & Spatial-Up       & 18.91  & 0.838   \\
                             & FourierUp-AreaUp       & \underline{19.16} & \underline{0.843}    \\
                             & FourierUp-Padding     & \textbf{19.35}  & \textbf{0.847}    \\\hline
    \multirow{4}{*}{MSBDN}   & Original        & 33.79  & 0.984    \\
                             & Spatial-Up       & 33.90  & 0.984    \\
                             & FourierUp-AreaUp       & \underline{34.21}  & \underline{0.985}    \\
                             & FourierUp-Padding     & \textbf{34.35}  & \textbf{0.987}    \\\hline
\end{tabular}
\label{tab:dh}
\end{minipage}
\hspace{0.0001\linewidth}
\begin{minipage}[c]{0.48\textwidth}
\small
\centering
\renewcommand{\arraystretch}{1.1}
\captionof{table}{\textbf{Comparison over object detection.}}
\begin{tabular}{c  c |c c}
    \hline
    Model                        & Methods      & AP50   & mAP      \\  \hline
    \multirow{4}{*}{Faster RCNN} & Original         & 79.13  & 79.10    \\
                                 & Spatial-Up       & 79.14  & 79.10    \\
                                 & FourierUp-AreaUp       & \underline{79.16}  & \underline{79.13}    \\
                                 & FourierUp-Padding      & \textbf{79.19}  & \textbf{79.15}    \\\hline
      \multirow{4}{*}{YOLO-V3}   & Original         & 81.68  & 81.63    \\
                                 & Spatial-Up       & 81.68  & 81.63    \\
                                 & FourierUp-AreaUp      & \underline{81.70}  & \underline{81.65}    \\
                                 & FourierUp-Padding     & \textbf{81.72}  & \textbf{81.68}    \\\hline
\end{tabular}
\label{tab:od}
\end{minipage}

\begin{table*}[!htb]
\small
\centering
\renewcommand{\arraystretch}{1.1}
\caption{\textbf{Quantitative comparisons of medical image segmentation.}}
\begin{tabular}{l l |c c c c}
    \hline
    \multirow{2}{*}{Model} & \multirow{2}{*}{Configurations}& \multicolumn{2}{c}{synapse} & \multicolumn{2}{c}{CANDI} \\
    & &DSC $\uparrow$ & HD95 $\downarrow$ & DSC $\uparrow$ & HD95 $\downarrow$ \\
    \hline
    \multirow{4}{*}{U-Net} & Original& 76.85& 39.70       & 86.50 & 3.946  \\
                        &Spatial-Up & 76.94 & 38.59       & 86.59 & 3.826  \\
                         &FourierUp-AreaUp & \underline{77.25} & \underline{36.02}      & \underline{86.63} & \underline{3.751}  \\
                        &FourierUp-Padding & \textbf{77.37} & \textbf{35.86}      & \textbf{86.70} & \textbf{3.327}  \\
    \hline
    \multirow{4}{*}{Att-UNet} & Original& 77.77& 36.02     & 86.29& 5.601  \\
                            &Spatial-Up & 77.85 & 35.91    & 86.35 & 5.588  \\
                             &FourierUp-AreaUp & \underline{78.11} & \underline{34.54}    & \underline{86.50} & \underline{4.851}  \\
                            &FourierUp-Padding & \textbf{78.34} & \textbf{34.29}    & \textbf{86.64} & \textbf{4.833}  \\
    \hline 
\end{tabular}
\label{tab:ig}
\end{table*}


\begin{table*}[!htb]
\small
\centering
\renewcommand{\arraystretch}{1.1}
\caption{\textbf{Quantitative comparisons of  pan-sharpening.}}
\resizebox{0.96\linewidth}{!}{ 
\begin{tabular}{l l |c c c c  c c c c}
    \hline
    \multirow{2}{*}{Model} & \multirow{2}{*}{Configurations}& \multicolumn{4}{c}{WorldView-II} & \multicolumn{4}{c}{GaoFen2} \\
    & &PSNR$\uparrow$ & SSIM$\uparrow$&SAM$\downarrow$&ERGAS$\downarrow$ &PSNR$\uparrow$ & SSIM$\uparrow$&SAM$\downarrow$  &EGAS$\downarrow$ \\
    \hline
    \multirow{4}{*}{PANNET} & Original& 40.817 & 0.963 & 0.025 & 1.055         & 43.066 & 0.968 & 0.018 & 0.855  \\
                            & Spatial-Up& 40.988 & 0.963 & 0.025 & 1.031        & 43.897 & 0.973 & 0.018 & 0.737  \\
                            & FourierUp-AreaUp& \underline{41.167} & \underline{0.963} & \underline{0.024} & \underline{1.010}        & \underline{45.964} & \underline{0.979} & \underline{0.015} & \underline{0.653}  \\
                            &FourierUp-Padding & \textbf{41.288} & \textbf{0.965} & \textbf{0.024} & \textbf{1.007}       & \textbf{46.145} & \textbf{0.982} & \textbf{0.012} & \textbf{0.622}  \\
    \hline
    \multirow{4}{*}{DCFNET} &  Original& 40.276 & 0.968 & 0.028 & 1.051       & 42.986 & 0.967 & 0.019 & 0.858 \\
                            &Spatial-Up & 40.319 & 0.968 & 0.028 & \textbf{1.046}    & 43.157 & 0.970 & 0.017 & 0.850  \\
                             & FourierUp-AreaUp& \underline{40.484} & \underline{0.968} & \underline{0.025} & \underline{1.115}    & \underline{43.881} & \underline{0.979} & \underline{0.014} & \underline{0.829}  \\
                            & FourierUp-Padding& \textbf{40.546} & \textbf{0.968} & \textbf{0.025} & \underline{1.102}   & \textbf{44.153} & \textbf{0.981} & \textbf{0.014} & \textbf{0.765}  \\
    \hline
\end{tabular}}
\label{tab:ps}
\end{table*}

\begin{figure}[h!t]
	\centering
	\includegraphics[width=0.93\textwidth]{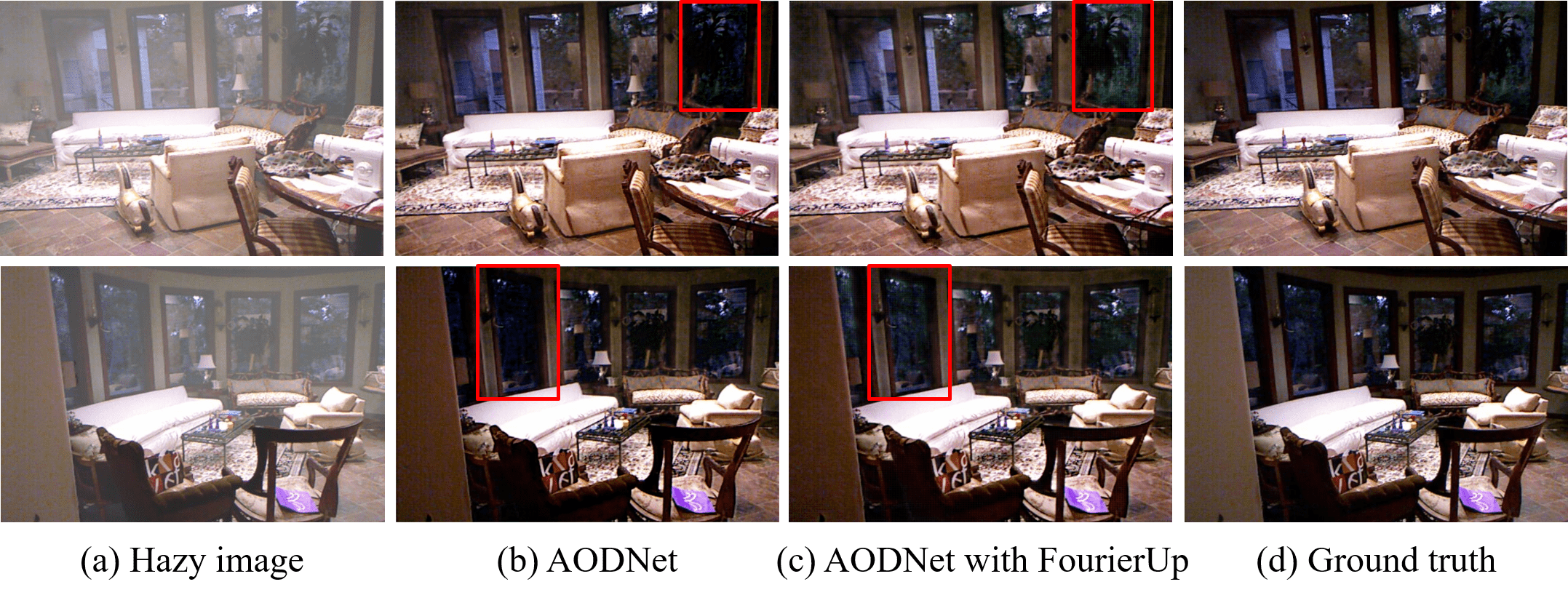}
 	\setlength{\abovecaptionskip}{-0.02cm}
	\caption{\textbf{Visual comparison of AODNet on the SOTS.}}
	\label{fig:dh}
  	\vspace{-2.6mm}
\end{figure}

\vspace{-3mm}
\section{Limitations} \label{seclimit}
\vspace{-2mm}
First, the more comprehensive experiments on broader computer vision  tasks (\emph{e.g.}, image de-noising and image de-blurring) have not been explored. Second, the deep Fourier Up-sampling integrated with spatial up-sampling will increase the model parameter numbers. This is negligible at the significant performance gain at fewer parameter increase.  Note that, the focus of this work is beyond designing a plug-and-play module to integrate it into existing networks for further performance gain. This work also provides a powerful scale change choice of the up-sampling operator pool when developing a new model from start.

\section{Conclusion}
\vspace{-2mm}
In this paper, we have proposed a  deep Fourier up-sampling to  explore the possibility of the up-sampling in the Fourier domain, which provides the key insight for the multi-scale Fourier pattern modeling. We theoretically demonstrate that our designs of Fourier up-sampling are feasible. It is appealing that the proposed FourierUp is a generic operator, thus being directly integrated with existing networks. Extensive experiments demonstrate the effectiveness of our method. We believe the FourierUp has the potential to advance  broader computer vision tasks, \emph{e.g.}, image/video super-resolution and image/video in-painting.

\vspace{-1mm}
\section*{Broader Impact} \label{broader}
\vspace{-1.2mm}
Our work shows the promising capability of up-sampling in the Fourier domain for computer vision algorithms through two novel designs with theoretical proofs. Using our deep Fourier up-sampling with negligible computational cost will improve the performance of neural networks and facilitate the development of AI in real-world applications. However, the efficacy of our method may raise  potential concerns when it is improperly used. For example, the safety of the applications of our method in real-world applications may not be guaranteed. We will investigate the robustness and effectiveness of our method in broader real-world applications.

\vspace{-1mm}
\section*{Acknowledgements} \label{ack}
\vspace{-1.2mm}
We gratefully acknowledge the support of the Major Key Project of PCL (PCL2021A12), MindSpore, CANN, and Ascend AI Processor used for this research. Chongyi Li and Chen Change Loy are supported by the RIE2020 Industry Alignment Fund – Industry Collaboration Projects (IAF-ICP) Funding Initiative, as well as cash and in-kind contribution from the industry partner(s).

\newpage
\section*{References}
\begingroup
\renewcommand{\section}[2]{}%
\bibliographystyle{unsrt}
\bibliography{ref.bib}
\endgroup

\section*{Checklist}

\begin{enumerate}

\item For all authors...
\begin{enumerate}
  \item Do the main claims made in the abstract and introduction accurately reflect the paper's contributions and scope?
    \answerYes{}
  \item Did you describe the limitations of your work?
    \answerYes{}, {see Section \ref{seclimit}}.
  \item Did you discuss any potential negative societal impacts of your work?
    \answerYes{}, {see Section \ref{broader}}.
  \item Have you read the ethics review guidelines and ensured that your paper conforms to them?
    \answerYes{}
\end{enumerate}

\item If you are including theoretical results...
\begin{enumerate}
  \item Did you state the full set of assumptions of all theoretical results?
    \answerYes{}, {See Section \ref{pp} and Section \ref{uc}.}
        \item Did you include complete proofs of all theoretical results?
    \answerYes{}, {See Section \ref{pp} and Section \ref{uc}.}
\end{enumerate}

\item If you ran experiments...
\begin{enumerate}
  \item Did you include the code, data, and instructions needed to reproduce the main experimental results (either in the supplemental material or as a URL)?
    \answerYes{}
  \item Did you specify all the training details (e.g., data splits, hyperparameters, how they were chosen)?
    \answerYes{}
        \item Did you report error bars (e.g., with respect to the random seed after running experiments multiple times)?
    \answerNo{}, the effect of random seed could almost be negligible since we set the same initiation seed during experiments. Reproducibility can be guaranteed.
        \item Did you include the total amount of compute and the type of resources used (e.g., type of GPUs, internal cluster, or cloud provider)?
    \answerYes{}
\end{enumerate}

\item If you are using existing assets (e.g., code, data, models) or curating/releasing new assets...
\begin{enumerate}
  \item If your work uses existing assets, did you cite the creators?
    \answerYes{}
  \item Did you mention the license of the assets?
    \answerYes{}
  \item Did you include any new assets either in the supplemental material or as a URL?
   \answerYes{}
  \item Did you discuss whether and how consent was obtained from people whose data you're using/curating?
   \answerYes{}, the data creators claim that they allow and encourage the data for scientific research.
  \item Did you discuss whether the data you are using/curating contains personally identifiable information or offensive content?
    \answerNA{}, We train and test the framework using the publicly accessible datasets.
\end{enumerate}

\item If you used crowdsourcing or conducted research with human subjects...
\begin{enumerate}
  \item Did you include the full text of instructions given to participants and screenshots, if applicable?
    \answerNA{}
  \item Did you describe any potential participant risks, with links to Institutional Review Board (IRB) approvals, if applicable?
    \answerNA{}
  \item Did you include the estimated hourly wage paid to participants and the total amount spent on participant compensation?
    \answerNA{}
\end{enumerate}

\end{enumerate}

\end{document}


\maketitle
\footnotetext[1]{Man Zhou and Hu Yu contribute equally.}
\footnotetext[2]{Corresponding author: Chongyi Li.}

This supplementary document is organized as follows:

Section \ref{derivation} provides the alternative solution of Deep Fourier Up-Sampling variant: ``Corner Interpolation''.  Due to the page limits, we only  present two variants in main manuscript . Specifically, the theoretical evidences and module construction for FourierUp of ``Corner Interpolation Variant'' are reported in Section \ref{theevd} and Section \ref{fmd} respectively. The experimental results of "Corner Interpolation Variant" are presented in Section \ref{cintegration}.

Section \ref{imde} provides the implementation details as shown in Fig. \ref{fig:up} and  Fig. \ref{fig:wup}. To be specific, the configurations including ``Original'', ``FourierUp-AreaUp'', ``FourierUp-Padding'', and ``Spatial-Up'' in ``Implementation Details'' of main manuscript  are illustrated. 

Section \ref{moreexp} provides more quantitative and qualitative experimental results. 

Section \ref{fea} provides more visualization of feature maps between the baselines and the ones integrated with the proposed ``FourierUp''.

\begin{figure}[h!t]
	\centering
	\includegraphics[width=\textwidth]{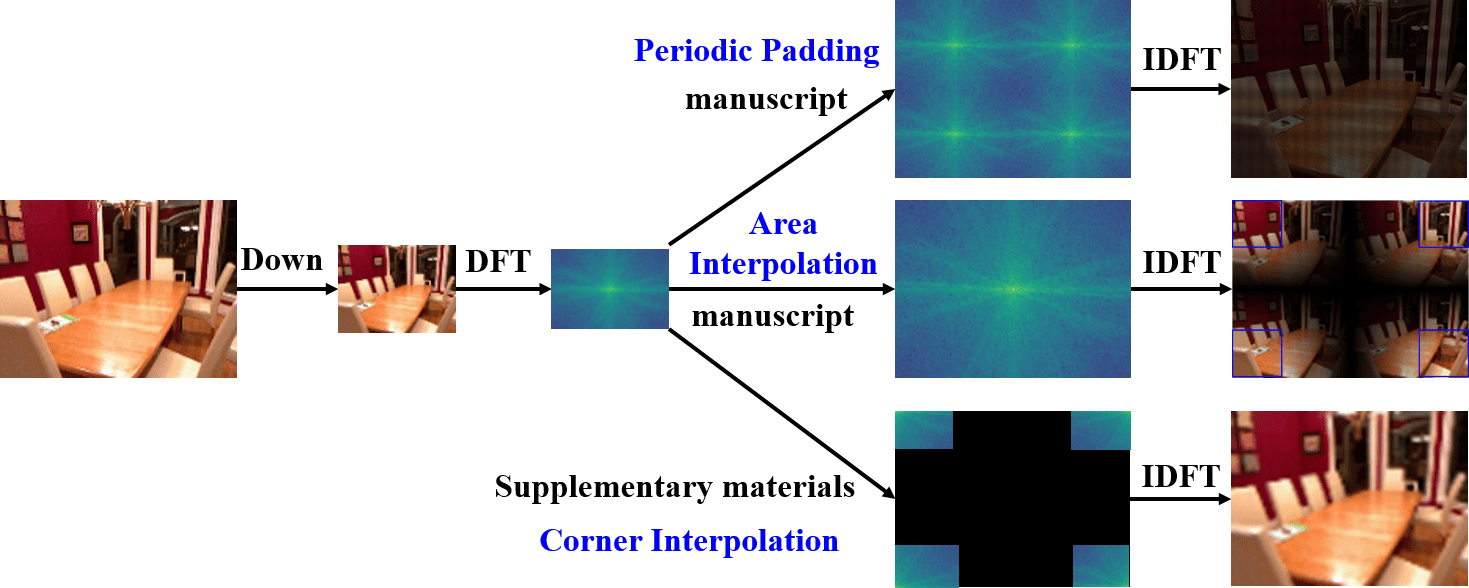}
	\caption{\textbf{An illustration of the proposed deep Fourier Up-Sampling.}  It has three alternative variants: Periodic Padding and Area Interpolation/Cropping, as illustrated in main manuscript . The alternative solution ``Corner Interpolation'' is shown in supplementary materials.}
	\label{fig:CIT}
\end{figure}

\section{Deep Fourier Up-Sampling Variant: ``Corner Interpolation''} \label{derivation}
We first illustrate the alternative solution ``Corner Interpolation'' of Deep Fourier Up-Sampling, and then present the theoretical evidences, finally detail the corresponding module construction in the peso-code. The illustration of ``Corner Interpolation'' is shown in Fig. \ref{fig:CIT}.  

\textbf{Theorem-1.} Suppose the shifted $F^{shift}_{G}$ of the Fourier map $G\in \mathbb{R}^{M\times N}$ as
\begin{equation}
	\small
		\begin{aligned}
          F^{shift}_{G}(u', v') = G(u-\frac{M}{2}, v-\frac{N}{2}),
  \end{aligned}
\end{equation}
where $u'=0,1,\dots,M-1$ and $v'=0,1,\dots,N-1$, it holds that the inverse Fourier transform $f^{shift}_{g}$ of $F^{shift}_{G}$
\begin{equation}
	\small
		\begin{aligned}
          f^{shift}_{g}(x, y) = (-1)^{(x+y)}g(x,y),
  \end{aligned}
\end{equation}
where $x=0,1,\dots,M-1$ and $y=0,1,\dots,N-1$.

\subsection{Theoretical Evidences for FourierUp of ``Corner Interpolation Variant''} \label{theevd}
For a spatial map $G(u,v) \in \mathbb{R}^{M\times N}$, we denote its $2$-times up-sampled corner interpolation as $F^{cor}_{G}(u, v) \in \mathbb{R}^{2M\times 2N}$. Denote $G(u,v)\in\mathbb{R}^{M\times N}$, $F^{cor}_{g}(u,v)\in\mathbb{R}^{2M\times 2N}$, $F^{shiftcor}_{g}(u,v)= F^{cor}_{G}(u-M, v-N)\in\mathbb{R}^{2M\times 2N}$ as the corresponding Fourier transforms of $g(x,y)$, $f^{cor}_{g}(x,y)$, and $f^{shiftcor}_{g}(x,y)$ respectively. 

The 2D Inverse Discrete Fourier transform (IDFT) of $G(u,v)$ can be written as
\begin{equation}
	\small
		\begin{aligned}
          g(x,y) = \frac{1}{MN}\sum_{u=0}^{M-1} \sum_{v=0}^{N-1} G(u,v) e^{j2\pi (\frac{ux}{M} + \frac{vy}{N})}.
  \end{aligned}
\end{equation}

We up-sample $G(u,v)\in \mathbb{R}^{M\times N}$ to get $F^{cor}_{G}(u,v) \in \mathbb{R}^{2M\times 2N}$ by corner interpolation. Specifically, the corner interpolation is shown in Fig. \ref{fig:subc}. For convenience, we infer the inverse Fourier transform of $F^{shiftcor}_{g}(u,v)$ as
\begin{equation} \label{hif}
	\resizebox{\textwidth}{!}{
\begin{math}
		\begin{aligned}
		  f^{shiftcor}_{G}(x,y) &= \frac{1}{4MN}\sum_{u=0}^{2M-1} \sum_{v=0}^{2N-1} F^{cor}_{G}(u,v) e^{j2\pi (\frac{ux}{2M} + \frac{vy}{2N})} \\
           &= \frac{1}{4MN} \sum_{u=0}^{M-1} \sum_{v=0}^{N-1} F^{AI}_{G}(2u, 2v) e^{j2\pi (\frac{2ux}{2M} + \frac{2vy}{2N})} + \frac{1}{4MN}
           \sum_{u=0}^{M-1} \sum_{v=0}^{N-1} F^{AI}_{G}(2u+1, 2v) e^{j2\pi (\frac{(2u+1)x}{2M} + \frac{2vy}{2N})}\\
           &= \frac{1}{4MN} \sum_{u=\frac{M}{2}}^{\frac{3M}{2}} \sum_{v=\frac{N}{2}}^{\frac{3N}{2}} F^{cor}_{G}(u,v) e^{j2\pi (\frac{2ux}{2M} + \frac{(2v+1)y}{2N})}\\
           &=\frac{1}{4MN} \sum_{u=\frac{M}{2}}^{\frac{3M}{2}} \sum_{v=\frac{N}{2}}^{\frac{3N}{2}} {G}(u-\frac{M}{2},v-\frac{N}{2}) e^{j2\pi (\frac{2ux}{2M} + \frac{(2v+1)y}{2N})}.
   \end{aligned}
\end{math}
}
\end{equation}
Let $u'=u-\frac{M}{2}$ and $v'=v-\frac{N}{2}$, the equation \eqref{hif} is transformed as
\begin{equation} \label{hift}
		\begin{aligned}
 f^{shiftcor}_{G}(x,y) &=\frac{1}{4MN} \sum_{u=\frac{M}{2}}^{\frac{3M}{2}} \sum_{v=\frac{N}{2}}^{\frac{3N}{2}} {G}(u-\frac{M}{2},v-\frac{N}{2}) e^{j2\pi (\frac{2ux}{2M} + \frac{(2v+1)y}{2N})}\\
 &= \frac{1}{4MN} \sum_{u'=0}^{M-1} \sum_{v'=0}^{N-1} G(u',v') e^{j2\pi (\frac{(u'+M/2)x}{2M} + \frac{(v'+N/2)y}{2N})}\\
 &= \frac{1}{4MN} \sum_{u'=0}^{M-1} \sum_{v'=0}^{N-1} G(u',v') e^{j2\pi (\frac{u(x/2)}{M} + \frac{v(y/2)}{N})} e^{j\pi (\frac{x}{2} + \frac{y}{2})}\\
  &=\frac{1}{4}g(\frac{x'}{2},\frac{y'}{2})e^{j\pi (\frac{x'}{2} + \frac{y'}{2})},
    \end{aligned}
\end{equation}
where $x'=2x$ and $y'=2y$, $x=0,1,\dots,M-1$ and $y=0,1,\dots,N-1$. Similarly, we can write $A(x,y)=e^{j\pi(\frac{x'}{2} +\frac{y'}{2})}$ as $|A(x,y)|=1$ for any integer $x$, $y$. Recall \textbf{Theorem-1}, we can infer $f^{cor}_{G}(x,y)$ as
\begin{equation} \label{hift}
		\begin{aligned}
 f^{cor}_{G}(x,y) &= (-1)^{(x+y)}f^{shiftcor}_{G}(x,y) \\
  &=g(\frac{x'}{2},\frac{y'}{2})e^{j\pi (\frac{x'}{2} + \frac{y'}{2})}\frac{(-1)^{(x+y)}}{4}.
    \end{aligned}
\end{equation}
We can find that the even points in $f^{cor}_{G}(2x, 2y)$ are equal to the corresponding point in $\frac{g(x, y)}{4}$. The odd points have no definitions and are obtained by interpolation, acting as the low-pass filtering in spatial domain. 

\textbf{Theorem-2.} Suppose the corner interpolated $F^{cor}_{G}$ of the Fourier map $G\in \mathbb{R}^{M\times N}$, it holds that the inverse Fourier transform $f^{cor}_{g}$ of $F^{cor}_{G}$
\begin{equation}
	\small
		\begin{aligned}
          f^{cor}_{g}(x, y) = g(\frac{x'}{2},\frac{y'}{2})e^{j\pi (\frac{x'}{2} + \frac{y'}{2})}\frac{(-1)^{(x+y)}}{4},
  \end{aligned}
\end{equation}
where $x'=2x$ and $y'=2y$, $x=0,1,\dots,M-1$ and $y=0,1,\dots,N-1$.

\begin{figure*}
\rule{\textwidth}{0.4pt}
    \begin{lstlisting}[language={Python}]
  def  FourierUp_CornerInterpolation(X):
 # X: input with shape [N, C, H, W]
 # A and P are the amplitude and phase
    A, P = FFT(X)

    # Fourier up-sampling transform rules
    A_aip = Corner-Interpolation(A)
    P_aip = Corner-Interpolation(P)
    A_aip = Convs_1x1(A_aip)
    P_aip = Convs_1x1(P_aip)

    # Inverse Fourier transform
    Y = iFFT(A_aip, P_aip)
    
    Return Y #[N, C, 2H, 2W]
\end{lstlisting}
    \hfill
    \vline
    \begin{lstlisting}[language={Python}]
def  Corner-Interpolation(X):
 # X: input with shape [N, C, H, W]
 # A and P are the amplitude and phase
    r, c = X.shape(2), X.shape(3)

    I_Mup=torch.zeros((N, C, 2*H, 2*W))
    I_Pup=torch.zeros((N, C, 2*H, 2*W))

    if r%2==1:#odd
        ir1,ir2=r//2+1,r//2+1
    else: #even
        ir1,ir2=r//2+1,r//2
    if c%2==1:#odd
        ic1,ic2=c//2+1,c//2+1
    else: #even
        ic1,ic2=c//2+1,c//2

    A_aip[:,:,:ir1,:ic1]=A[:,:,:ir1,:ic1]  
    A_aip[:,:,:ir1,ic2+c:]=A[:,:,:ir1,ic2:]
    A_aip[:,:,ir2+r:,:ic1]=A[:,:,ir2:,:ic1]
    A_aip[:,:,ir2+r:,ic2+c:]=A[:,:,ir2:,ic2:]

    if r%2==0:#even
        A_aip[:,:,ir2,:]=A_aip[:,:,ir2,:]*0.5
        A_aip[:,:,ir2+r,:]=A_aip[:,:,ir2+r,:]*0.5
    if c%2==0:#even
        A_aip[:,:,:,ic2]=A_aip[:,:,:,ic2]*0.5
        A_aip[:,:,:,ic2+c]=A_aip[:,:,:,ic2+c]*0.5


    P_aip[:,:,:ir1,:ic1]=P[:,:,:ir1,:ic1]  
    P_aip[:,:,:ir1,ic2+c:]=P[:,:,:ir1,ic2:]
    P_aip[:,:,ir2+r:,:ic1]=P[:,:,ir2:,:ic1]
    P_aip[:,:,ir2+r:,ic2+c:]=P[:,:,ir2:,ic2:]

    if r%2==0:#even
        I_Pup[:,:,ir2,:]=P_aip[:,:,ir2,:]*0.5
        I_Pup[:,:,ir2+r,:]=P_aip[:,:,ir2+r,:]*0.5
    if c%2==0:#even
        P_aip[:,:,:,ic2]=P_aip[:,:,:,ic2]*0.5
        P_aip[:,:,:,ic2+c]=P_aip[:,:,:,ic2+c]*0.5
    
    Return A_aip, P_aip
    \end{lstlisting}

\rule{\textwidth}{1pt}
\caption{\textbf{Pseudo-code of the variant of the proposed deep Fourier up-sampling: Corner interpolation variant.}}
    \label{fig:my_label}
\end{figure*}

\subsection{FourierUp Module Design of ``Corner Interpolation Variant''} \label{fmd}
Recalling \textbf{Theorem-1} and \textbf{Theorem-2}, we correspondingly propose the Fourier up-sampling module, called Corner Interpolation Variant.

\begin{figure}[h!t]
\vspace{-2em}
	\centering
	\includegraphics[width=0.7\textwidth]{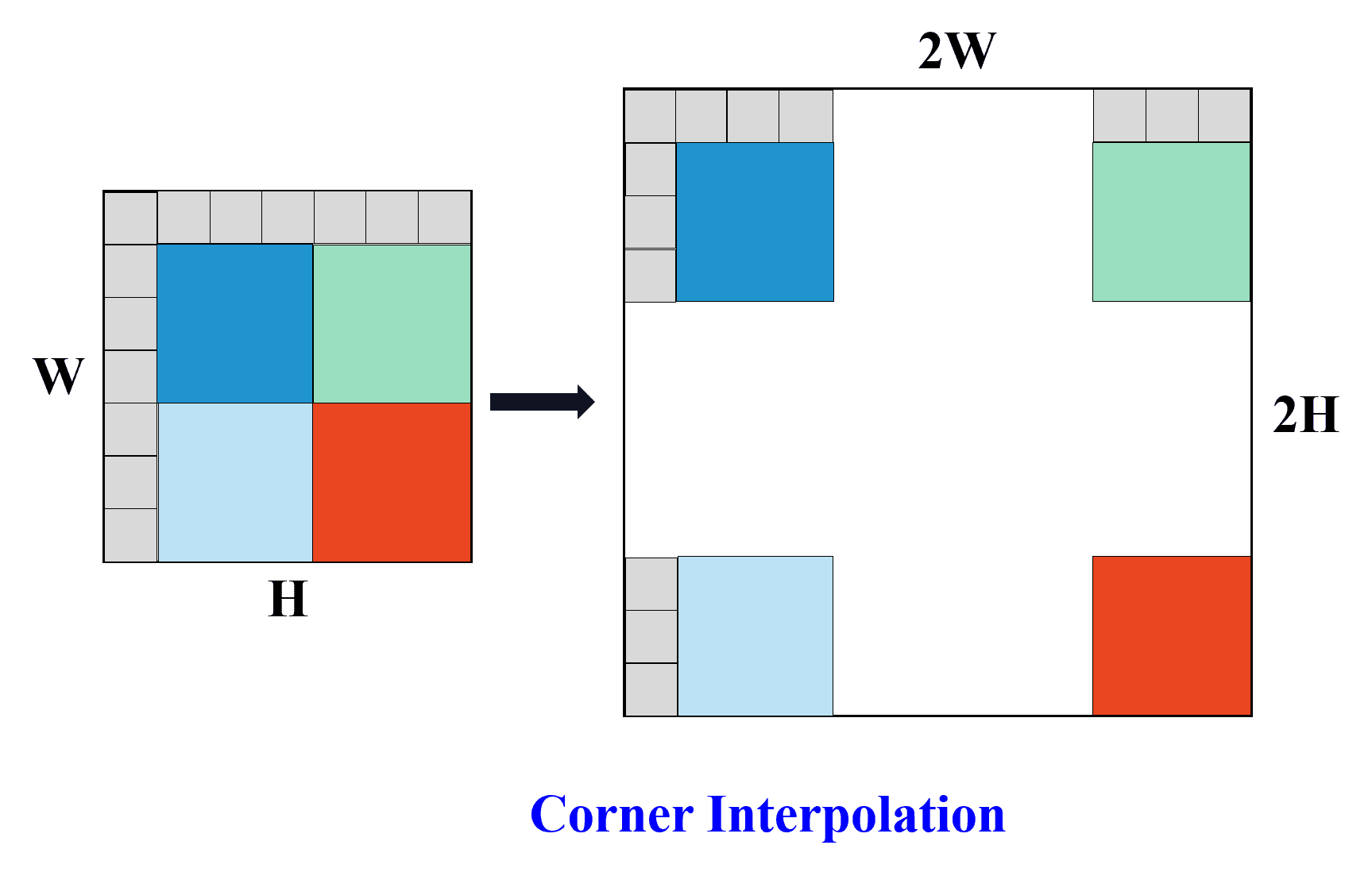}
	\caption{\textbf{The illustrations of corner interpolation implementation in Fig. \ref{fig:CIT}.} The gray parts represent a row/column pixels while the remaining color parts are evenly divided.}
	\label{fig:subc}
\end{figure}

\textbf{Corner Interpolation Up-Sampling.} The pseudo-code of the Corner Interpolation Up-sampling is shown in Fig. \ref{fig:my_label}. Specifically, given an image $\mathrm{X \in \mathbb{R}^{H \times W \times C}}$, we first adopt the Fourier transform $\mathrm{FFT(X)}$ to obtain its amplitude component $\mathrm{A}$ and phase component $\mathrm{P}$. We then perform the Corner Interpolation over $\mathrm{A}$ and $\mathrm{P}$  two times in both the $H$ and $W$ dimensions, as illustrated in Fig. \ref{fig:subc}, to form the padded $\mathrm{A\_pep}$ and $\mathrm{P\_pep}$. Such up-sampling maps are then fed into two independent convolution modules with a kernel size of $1\times 1$ and followed by the inverse Fourier transform $\mathrm{iFFT(.)}$ to project the padded ones back to the spatial domain.

Note that albeit being designed on the basis of strict theories, both constructed spectral up-sampling modules contain certain approximations, like a learnable $1\times 1$ convolution operator instead of strictly $1/4$ as described in Theorem-1 of main manuscript , and an approximation cropping to preserve the map corners instead of accurate $A$ mapping as proved in Theorem-2 of main manuscript  and Theorem-2 of supplementary materials. Such strategy makes the proposed modules able to be more easily implemented and more flexibly represent real data spectral structures. It is worth noting that this should be the first attempt for constructing easy equitable spectral upsampling modules, and hope it would inspire more effective and rational ones from more spectral perspectives.

\begin{table*}[!htb]
\small
\centering
\renewcommand{\arraystretch}{1.0}
\caption{\textbf{Quantitative comparisons of image de-raining.}}
\begin{tabular}{l l |c c c c }
    \hline
    \multirow{2}{*}{Model} & \multirow{2}{*}{Configurations}& \multicolumn{2}{c}{Rain200H} & \multicolumn{2}{c}{Rain200L}  \\
    &&PSNR & SSIM & PSNR & SSIM  \\
    \hline
    \multirow{4}{*}{LPNet} & Original& 22.907 & 0.775 & 32.461& 0.947  \\
     &Spatial-Up & 22.956 & 0.777 & 32.522 & 0.950  \\
    &FourierUp-AreaUp & 22.163 & 0.783 & 32.681 & 0.954  \\
    &FourierUp-Corner & \underline{22.291} & \underline{0.786} & \underline{32.678} & \underline{0.954}  \\
    &FourierUp-Padding & \textbf{23.295} & \textbf{0.786} & \textbf{32.835} & \textbf{0.956}  \\
    \hline
    \multirow{4}{*}{PReNet} & Original& 29.041
 & 0.891 & 37.802  & 0.981  \\
     &Spatial-Up & 29.357 & 0.901 & 38.271 & 0.985  \\
    &FourierUp-AreaUp & 29.690 & 0.903 & 39.776 & 0.985 \\
    &FourierUp-Corner & \underline{29.866} & \underline{0.908} & \underline{39.970} & \underline{0.987} \\
    &FourierUp-Padding & \textbf{29.871} & \textbf{0.908} & \textbf{39.971}  & \textbf{0.987}  \\
    \hline
\end{tabular}
\label{tab:dr}
\vspace{-1em}
\end{table*}

\begin{table*}[!htb]
\small
\centering
\renewcommand{\arraystretch}{1.4}
\caption{\textbf{Quantitative comparisons of  pan-sharpening.}}
\resizebox{\textwidth}{!}{ 
\begin{tabular}{l l |c c c c  c c c c}
    \hline
    \multirow{2}{*}{Model} & \multirow{2}{*}{Configurations}& \multicolumn{4}{c}{WorldView-II} & \multicolumn{4}{c}{GaoFen2} \\
    & &PSNR$\uparrow$ & SSIM$\uparrow$&SAM$\downarrow$&ERGAS$\downarrow$ &PSNR$\uparrow$ & SSIM$\uparrow$&SAM$\downarrow$  &EGAS$\downarrow$ \\
    \hline
    \multirow{4}{*}{PANNET} & Original& 40.817 & 0.963 & 0.025 & 1.055         & 43.066 & 0.968 & 0.018 & 0.855  \\
                            & Spatial-Up& 40.988 & 0.963 & 0.025 & 1.031        & 43.897 & 0.973 & 0.018 & 0.737  \\
                            & FourierUp-AreaUp& 41.167 & 0.963 & 0.024 & 1.010        & 45.964 & 0.979 & 0.015 & 0.653  \\
                              & FourierUp-Corner& \underline{41.286} & \underline{0.965} & \underline{0.024} & \underline{1.007}        & \underline{46.137} & \underline{0.981} & \underline{0.012} & \underline{0.631}  \\
                            &FourierUp-Padding & \textbf{41.288} & \textbf{0.965} & \textbf{0.024} & \textbf{1.007}       & \textbf{46.145} & \textbf{0.982} & \textbf{0.012} & \textbf{0.622}  \\
    \hline
    \multirow{4}{*}{DCFNET} &  Original& 40.276 & 0.968 & 0.028 & 1.051       & 42.986 & 0.967 & 0.019 & 0.858 \\
                            &Spatial-Up & 40.319 & 0.968 & 0.028 & \textbf{1.046}    & 43.157 & 0.970 & 0.017 & 0.850  \\
                             & FourierUp-AreaUp& 40.484 & 0.968 & 0.025 & 1.115    & 43.881 & 0.979 & 0.014 & 0.829  \\
                             & FourierUp-Corner& \underline{40.539} & \underline{0.968} & \underline{0.027} & 1.105    & \underline{44.139} & \underline{0.979} & \underline{0.014} & \underline{0.771}  \\
                            & FourierUp-Padding& \textbf{40.546} & \textbf{0.968} & \textbf{0.025} & \underline{1.102}   & \textbf{44.153} & \textbf{0.981} & \textbf{0.014} & \textbf{0.765}  \\
    \hline
\end{tabular}}
\label{tab:ps}
\end{table*}

\subsection{Comparison and Analysis} \label{cintegration}
 We perform the model performance comparison over different configurations, as described in implementation details of main manuscript . The quantitative results are presented in Tables \ref{tab:dr} to \ref{tab:ps} where the best and second best results are highlighted in bold and underline. From the results, by integrating with the FourierUp variant ``Corner-interpolation'', we can observe performance gain against the baselines across all the datasets in two representative tasks: image de-raining and pan-sharpening, suggesting the effectiveness of our approach. For example, for the PReNet of Table \ref{tab:dr}, ``FourierUp-Corner'' obtains 0.82dB/2.1dB  PSNR gains than the  ``Original'', 0.51dB/1.5dB PSNR gains than ``Spatial-Up'' on the Rain200H and Rain200L datasets, respectively. Such results validate the effectiveness of our proposed FourierUp.

\begin{figure}[h!t]
	\centering
	\includegraphics[width=\textwidth]{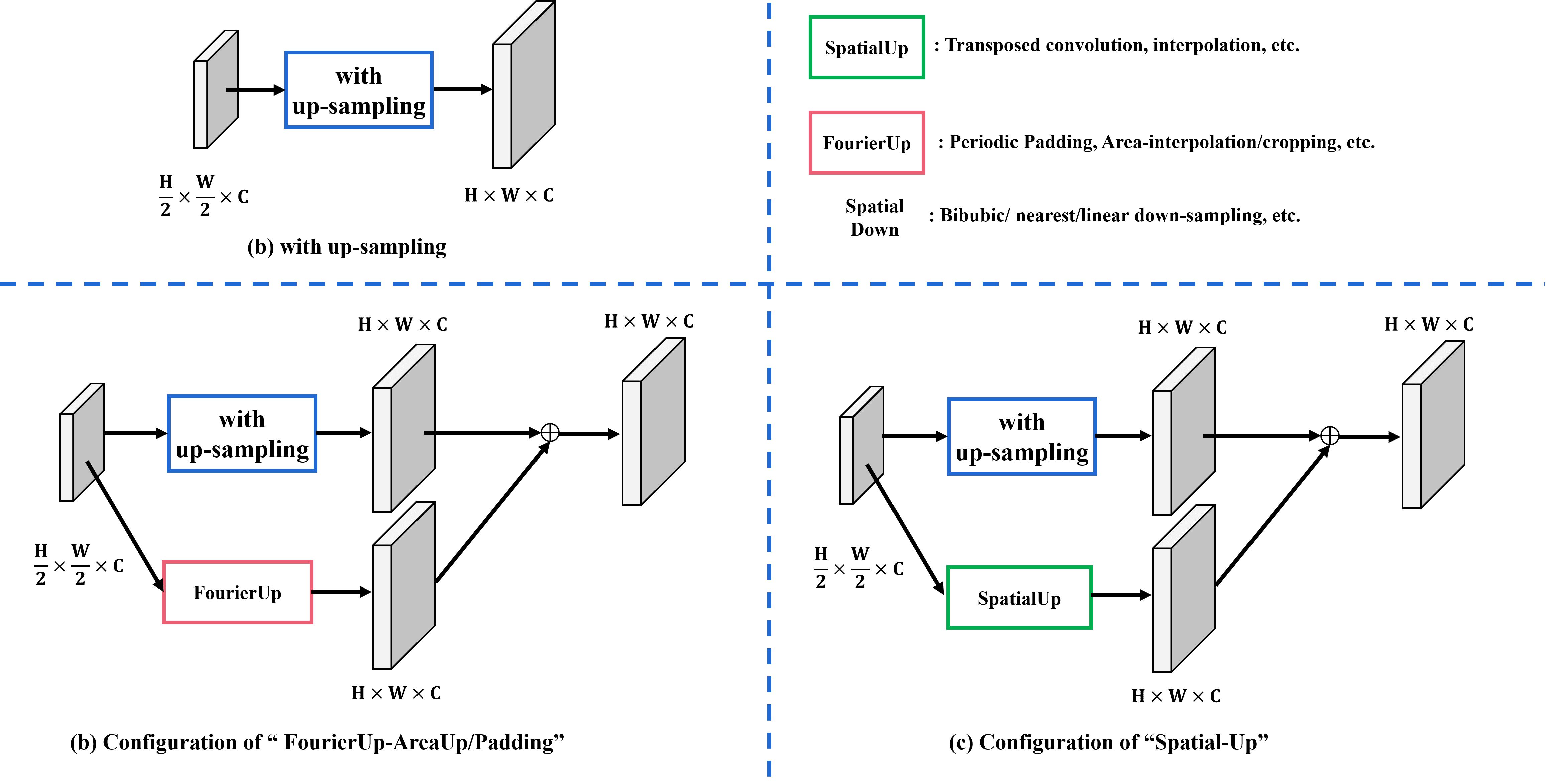}
	\caption{\textbf{The illustrations of corner interpolation over the baselines with up-sampling in Figure \ref{fig:CIT}.} The gray parts represent a row/column pixels while the remaining color parts are evenly divided.}
	\label{fig:up}
\end{figure}

\begin{figure}[h!t]
	\centering
	\includegraphics[width=\textwidth]{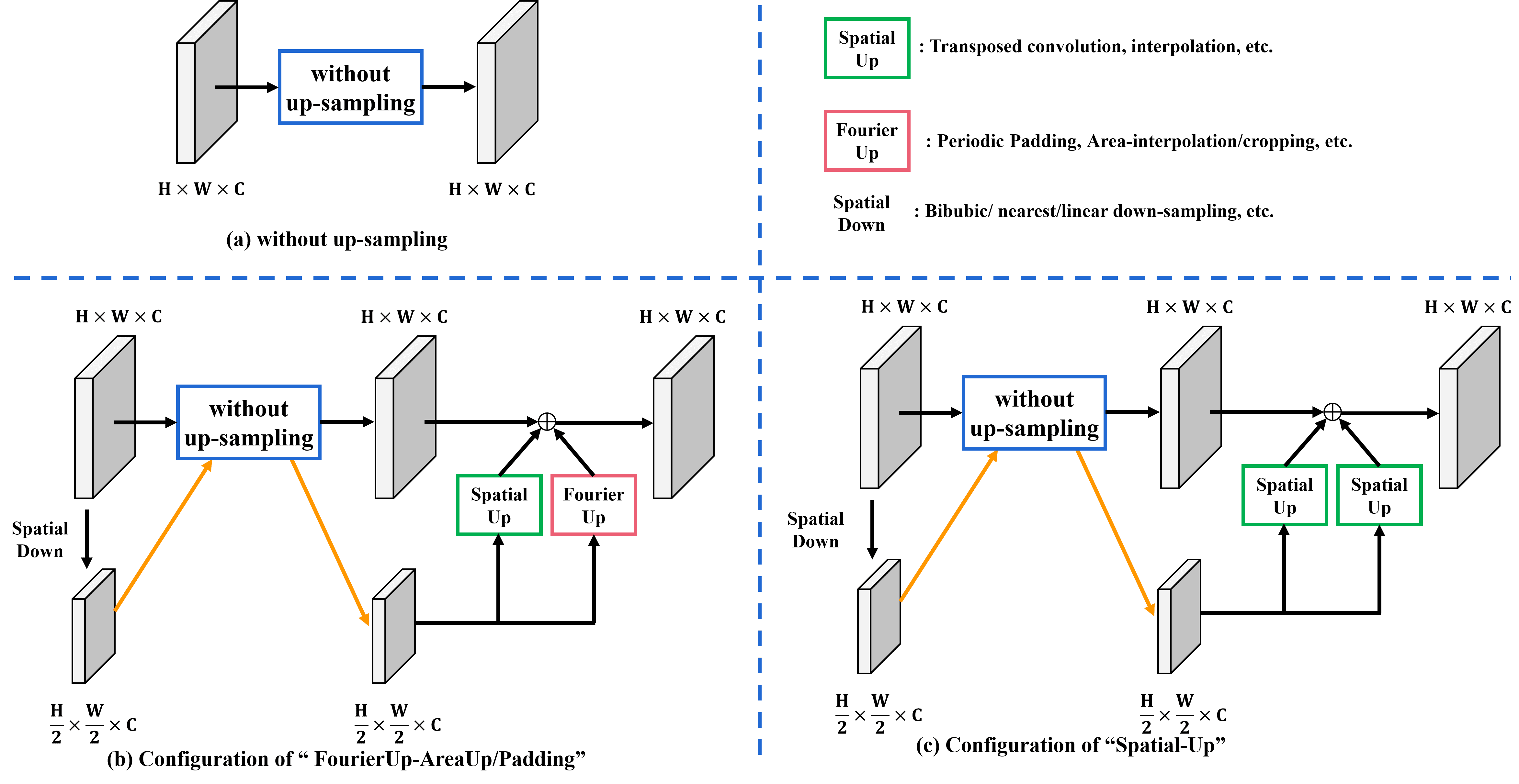}
	\caption{\textbf{The illustrations of corner interpolation over the baselines without up-sampling in Figure \ref{fig:CIT}.} The gray parts represent a row/column pixels while the remaining color parts are evenly divided.}
	\label{fig:wup}
	\vspace{-2em}
\end{figure}

\section{Implementation Details} \label{imde}
Regarding the above competitive baselines, they can be divided into two categories: one with spatial up-sampling and another one without spatial up-sampling. We provide the detailed structures of the encapsulated FourierUp (detailed in Fig. \ref{fig:wup}(b)) and the baselines with the FourierUp in Fig. \ref{fig:wup} and Fig. \ref{fig:up}.

For the baselines with spatial up-sampling, we perform the comparison over four configurations:
\begin{itemize}
\setlength{\itemsep}{0pt}
\setlength{\parsep}{0pt}
\setlength{\parskip}{0pt}
\item [1)] \textbf{Original}: the baseline without any changes; 
\item [2)] \textbf{FourierUp-AreaUp in Fig. \ref{fig:up}(b)}:  replacing the original model's spatial up-sampling  with the union of the Area-Interpolation variant of our FourierUp and the spatial up-sampling itself;
\item[3)] \textbf{FourierUp-Padding in Fig. \ref{fig:up}(b)}:  replacing the original model's spatial up-sampling operator  with the union of the Periodic-Padding variant of our FourierUp and  the spatial up-sampling itself;
\item[4)] \textbf{FourierUp-Corner in Fig. \ref{fig:up}(b)}:  replacing the original model's spatial up-sampling operator  with the union of the Corner-Interpolation variant of our FourierUp and  the spatial up-sampling itself;
\item[5)] \textbf{Spatial-Up in Fig. \ref{fig:up}(c)}:   replacing the variants of  FourierUp in the settings of $2)/3)$ with the spatial up-sampling. For a  fair comparison, we use the same number of trainable parameters as $2)/3)$.
\end{itemize}

For the baselines without spatial up-sampling, we perform the comparison over four configurations:
\begin{itemize}
\setlength{\itemsep}{0pt}
\setlength{\parsep}{0pt}
\setlength{\parskip}{0pt}
\item [1)] \textbf{Original}: the baseline without any changes; 
\item [2)] \textbf{FourierUp-AreaUp in Fig. \ref{fig:wup}(b)}:  replacing the original model's convolution with the encapsulated FourierUp that is equipped with the Area-Interpolation variant;
\item[3)] \textbf{FourierUp-Padding in Fig. \ref{fig:wup}(b)}: replacing the original model's convolution  with the encapsulated FourierUp that is equipped with the Periodic-Padding variant;
\item[4)] \textbf{FourierUp-Corner in Fig. \ref{fig:wup}(b)}: replacing the original model's convolution  with the encapsulated FourierUp that is equipped with the Corner-Interpolation variant;
\item[5)] \textbf{Spatial-Up in Fig. \ref{fig:wup}(c)}:   replacing the the encapsulated FourierUp of the settings of $2)/3)$ with the spatial up-sampling. For a  fair comparison, we use the same number of trainable parameters as $2)/3)$.
\end{itemize}

\section{Experiments} \label{moreexp}
\textbf{Quantitative Comparison.} We adopt the pan-sharpening, the representative task of guided image super-resolution, for evaluations. Due to the page limits, the results over WorldView III have not been presented in main manuscript.  We also employ  two different network designs for validation, including PANNET without up-sampling operator and DCFNET with up-sampling operator.

\begin{figure}[h!t]
	\centering
	\includegraphics[width=\textwidth]{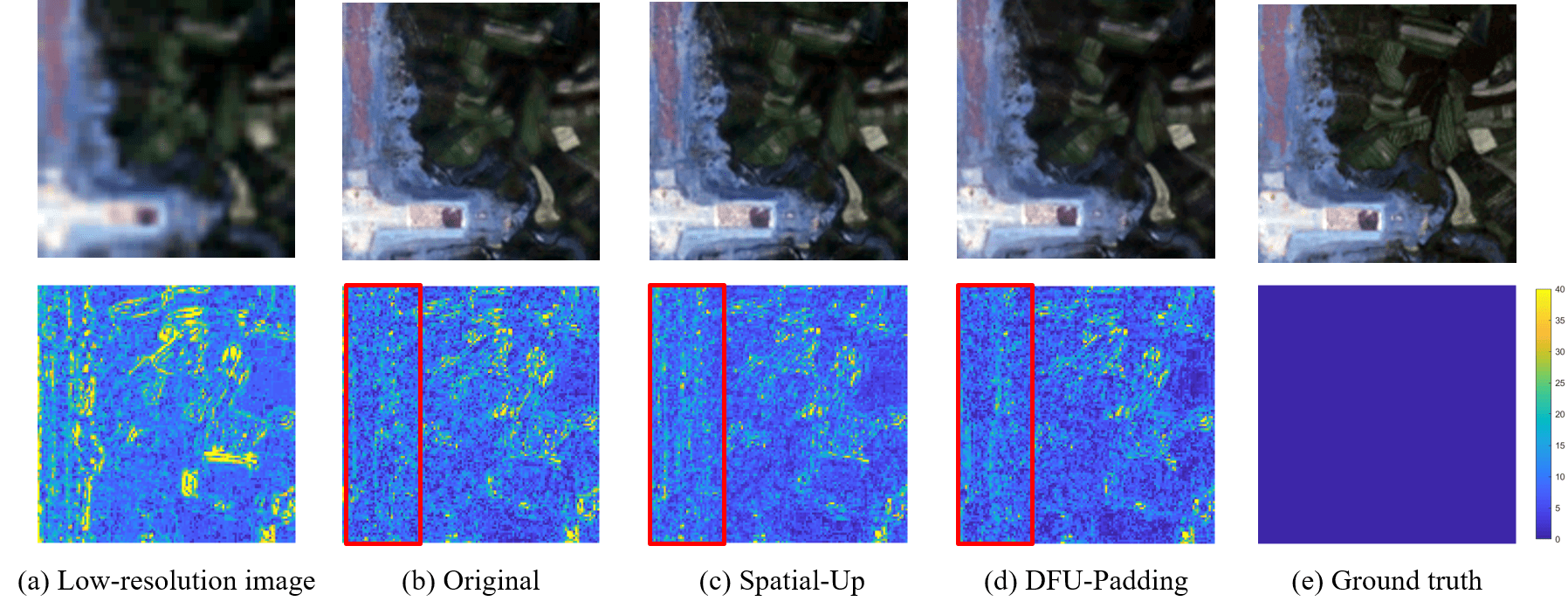}
	\caption{The visual comparison of PANNET over WorldView-II.}
	\label{fig:ps}
\end{figure}

We perform the model performance comparison over different configurations, as described in implementation details. The quantitative results are presented in Table \ref{tab:cmps} where the best and second best results are highlighted in bold and underline. From the results, by integrating with our proposed FourierUp variants, we can observe performance gain against the baselines across all the datasets, suggesting the effectiveness of our approach. For example, for the PANNET of Table \ref{tab:cmps}, ``FourierUp-padding'',  ``FourierUp-AreaUp'' and ``FourierUp-Corner''  obtain the  performance gains than the  ``Original'' and ``Spatial-Up'' on the WorldView-III datasets, respectively. Such results validate the effectiveness of our proposed FourierUp.  The corresponding visualization consistently supports the analysis in Fig. \ref{fig:ps}, where the FourierUp is capable of better maintaining the details.

\begin{figure}[h!t]
	\centering
	\includegraphics[width=\textwidth]{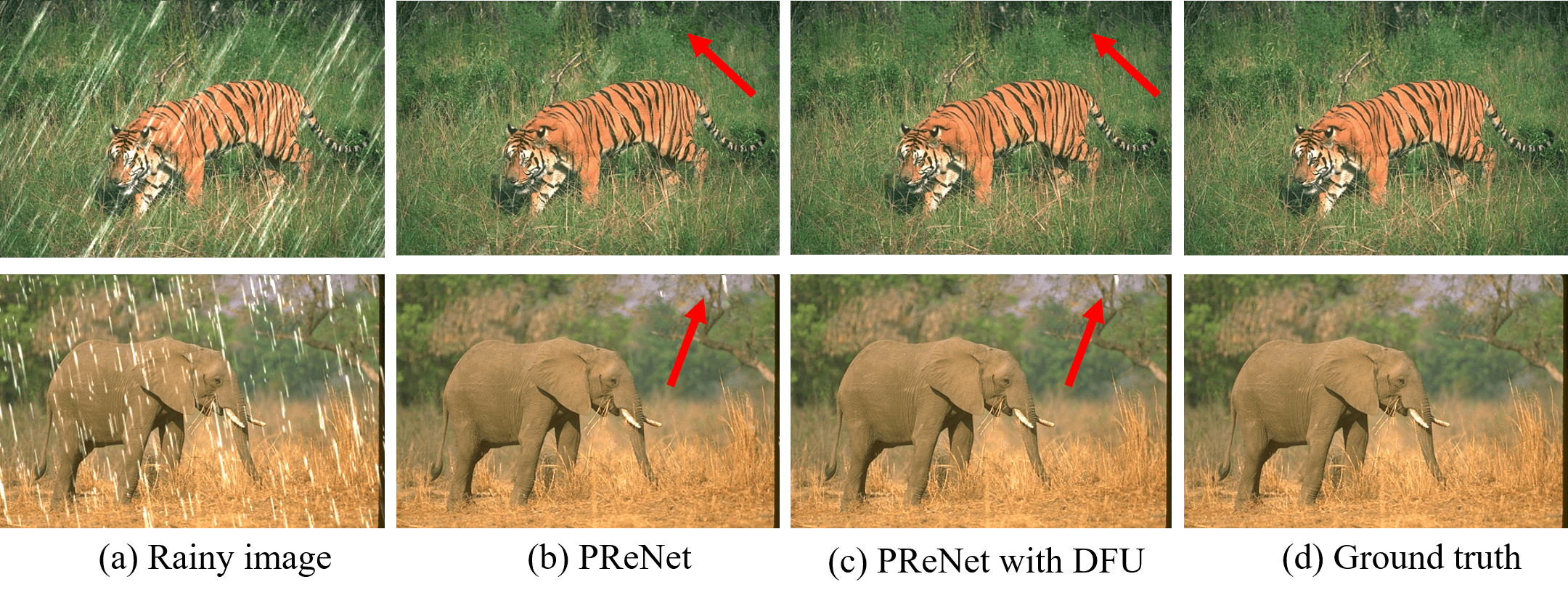}
 	\setlength{\abovecaptionskip}{-0.02cm}
	\caption{\textbf{Visual comparison of PReNet on the Rain200L.}}
	\label{fig:dr}
\end{figure}

\begin{figure}[h!t]
	\centering
	\includegraphics[width=\textwidth]{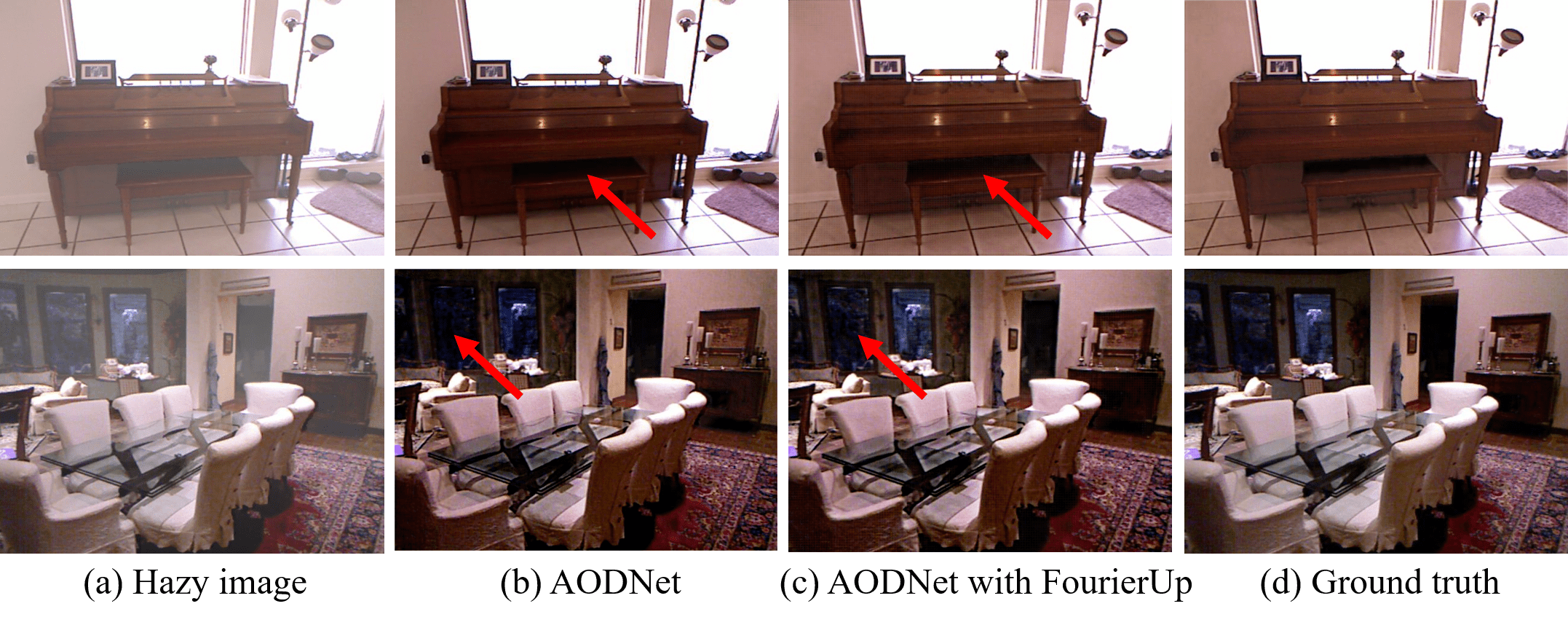}
 	\setlength{\abovecaptionskip}{-0.02cm}
	\caption{\textbf{Visual comparison of AODNet on the SOTS.}}
	\label{fig:dh}
\end{figure}

\textbf{Qualitative  Comparison.} Due to the limited space, we only report the visual results of the de-raining/dehazing task in main manuscript. We report more visual results  in the supplementary materials. As shown, integrating the FourierUp with the original baseline achieves more visually pleasing results. Specifically, zooming-in the red box arrows of Fig.  \ref{fig:dr} and \ref{fig:dh}, the model equipped with the FourierUp is capable of better recovering the texture details while removing the rain/haze effect.

\begin{table*}[!htb]
\small
\centering
\renewcommand{\arraystretch}{1.3}
\caption{\textbf{Quantitative comparisons of  pan-sharpening.}}
\begin{tabular}{l l |c c c c}
    \hline
    \multirow{2}{*}{Model} & \multirow{2}{*}{Configurations}& \multicolumn{4}{c}{WorldView-III}  \\
    & &PSNR$\uparrow$ & SSIM$\uparrow$&SAM$\downarrow$&ERGAS$\downarrow$ \\
    \hline
    \multirow{4}{*}{PANNET} & Original& 29.68 & 0.907 & 0.085 & 3.426      \\
                            & Spatial-Up& 29.71 & 0.907 & 0.085 & 3.426       \\
                            & FourierUp-AreaUp& \underline{30.05} & \underline{0.915} & \underline{0.078} & \underline{3.253}         \\
                            &FourierUp-Padding & \textbf{30.24} & \textbf{0.918} & \textbf{0.077} & \textbf{3.187}        \\
    \hline
    \multirow{4}{*}{DCFNET} &  Original& 29.47 & 0.907 & 0.089 & 3.536     \\
                            &Spatial-Up & 29.51 & 0.907 & 0.088 & 3.536      \\
                             & FourierUp-AreaUp& \underline{29.69} & \underline{0.913} & \underline{0.085} & \underline{3.326}      \\
                            & FourierUp-Padding& \textbf{29.85} & \textbf{0.914} & \textbf{0.078} & \underline{3.219}     \\
    \hline
\end{tabular}
\label{tab:cmps}
\vspace{-2mm}
\end{table*}

\section{Comparison in Feature Space} \label{fea}
\begin{figure}[h!t]
	\centering
	\includegraphics[width=\textwidth]{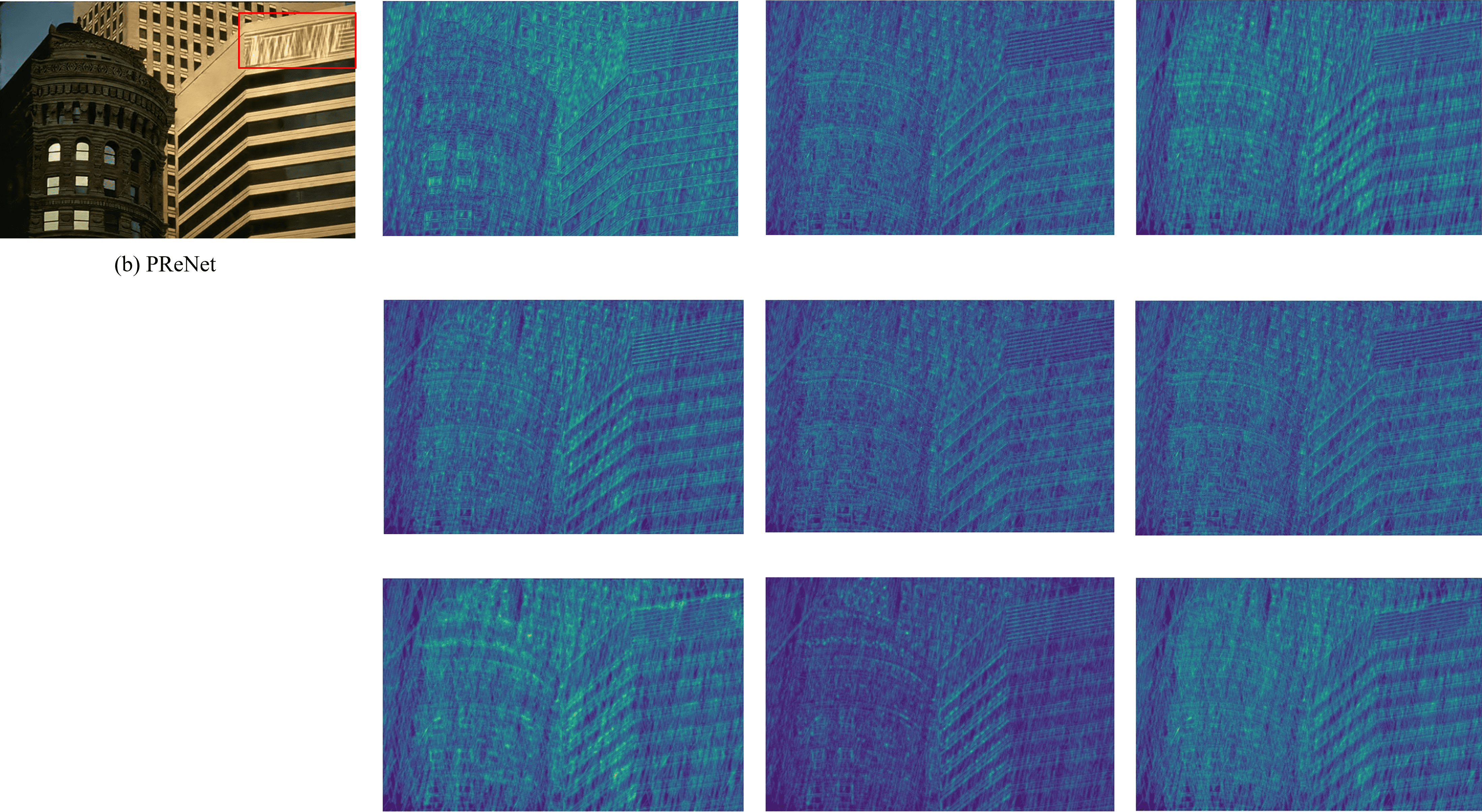}
	\caption{The visual feature maps comparison.}
	\label{fig:features}
\end{figure}

\begin{figure}[h!t]
	\centering
	\includegraphics[width=\textwidth]{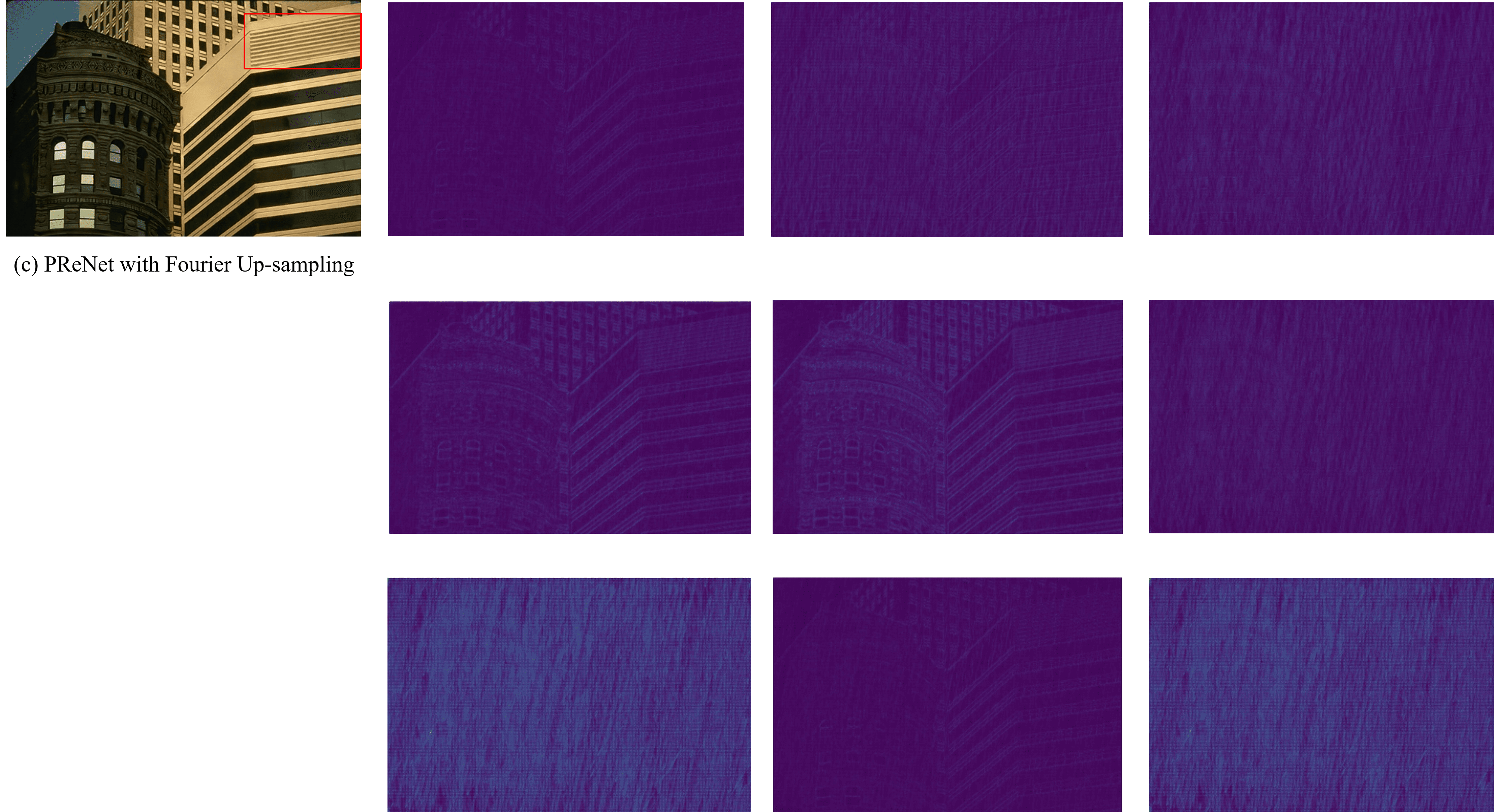}
	\caption{The visual feature maps comparison.}
	\label{fig:feature}
\end{figure}

In this section, we present more visualization results of feature maps, demonstrating the effectiveness of the FourierUp. Fig. \ref{fig:features} and Fig. \ref{fig:feature} present the representative example in the PreNet over image de-raining dataset-Rain200H. As can be seen, the PreNet integrated with our proposed FourierUp is capable of better distinguishing and disentangling the rain features and background features, thus improving the model performance while the original PreNet suffers from severe feature entanglement over rain streaks and background.

